\documentclass[1p,12pt]{elsarticle}
\usepackage[T1]{fontenc}
\usepackage[utf8]{inputenc}
\usepackage{lmodern}
\usepackage{amsmath,amssymb,booktabs,graphicx,array}
\usepackage{pdflscape} 
\usepackage{soul}
\usepackage{pgfplots}
\pgfplotsset{compat=1.18}
\usetikzlibrary{arrows.meta,positioning,calc}
\usepackage[hidelinks]{hyperref}

\newcommand{\captionmodel}{Caption + text}
\makeatletter
\def\ps@pprintTitle{\let\@oddhead\@empty\let\@evenhead\@empty\def\@oddfoot{\hfil\thepage\hfil}\let\@evenfoot\@oddfoot}
\makeatother

\begin{document}

\begin{frontmatter}

\title{Image Clustering with Vision-Language Model Embeddings}
\author[lasige]{Jo\~ao P. Matos-Carvalho}
\author[copelabs,cts]{Nuno Fachada}
\affiliation[lasige]{organization={LASIGE, Faculdade de Ci\^encias, Universidade de Lisboa},
 city={Lisboa}, postcode={1749-016}, country={Portugal}}
\affiliation[copelabs]{organization={COPELABS, Lus\'ofona University},
 city={Lisboa}, postcode={1700-921}, country={Portugal}}
\affiliation[cts]{organization={INESC INOV},
 city={Lisboa}, postcode={1000-029}, country={Portugal}}

\begin{abstract}
Pretrained vision--language models provide image representations whose suitability for clustering may vary with feature extraction, image domain, and partitioning algorithm. We compare nine representation pipelines across seven image benchmarks and four clustering configurations, yielding 252 baseline combinations. Eight pipelines extract visual features, including two vision-only references; the ninth generates descriptions and embeds the resulting text. With agglomerative clustering (Ward linkage) fixed, Qwen3-VL achieves the highest adjusted Rand index (ARI) on five datasets, including 0.925 on MNIST, while BLIP-2 leads on CIFAR-10 and CIFAR-100 with ARIs of 0.925 and 0.591, respectively. A separate sensitivity analysis compares raw and L2-normalized embeddings at their original dimensionality across all 63 dataset--representation pairs. Normalization preserves the highest-ARI representation in every dataset, increases ARI in 35 pairs, decreases it in 26, and leaves two numerically unchanged. On the SVHN dataset, the highest ARI remains low, rising from 0.097 to 0.098 after normalization. The evaluated caption-mediated pipeline obtains relatively high ARI on isolated digits but does not consistently exceed direct visual features. These findings support selecting and evaluating complete representation--clustering pipelines for the intended domain, with explicit attention to extraction and normalization choices.
\end{abstract}

\begin{keyword}
image clustering \sep vision--language models \sep image embeddings \sep multimodal representation learning \sep clustering evaluation
\end{keyword}

\end{frontmatter}

\section{Introduction}
Image clustering organizes an unlabeled collection by grouping images according to a representation of their content. This enables exploratory analysis without fitting a supervised classifier to the target categories. However, the category structure recovered by a clustering algorithm depends on the information emphasized by the representation. Images may be similar in background, shape, or color while depicting different objects, and the same collection can support several meaningful partitions. Research on semantic image clustering and text-guided alternative clustering illustrates these distinct aspects of the problem~\cite{scan,tgaicc}.

Pretrained vision--language models (VLMs) provide representations that can be reused across image domains. CLIP learns from paired images and text~\cite{clip}, while SigLIP2 combines language supervision with additional objectives for visual representation learning~\cite{siglip2}. Generative systems, including BLIP-2, LLaVA, and Qwen3-VL, also make visual features and image descriptions accessible through different components~\cite{blip2,llava,qwen}. These models support distinct representation routes, including visual feature aggregation and encoding generated descriptions. The resulting vectors differ in the information available to downstream clustering.

Recent work has explicitly investigated these alternatives. Stephan et al. use image captioning and visual question answering to produce text representations for clustering, and show that prompts can introduce task or domain knowledge~\cite{tgic}. Instead, VLM2Vec and mmE5 train models to produce multimodal embeddings, emphasizing that embedding quality can be an explicit learning objective~\cite{vlm2vec,mme5}. Other approaches adapt the visual representation to clustering through cross-modal consistency or combine generative guidance with ensembles~\cite{crossmodal2025,gsec2026}. These studies illustrate the different ways in which language and visual representations can be used to support image clustering.

A complementary empirical question concerns the behavior of available representations when no additional representation training is performed on the target datasets. This setting is relevant when computational or data constraints preclude model adaptation. Feature extraction and the clustering procedure then determine how pretrained representations are used. Comparisons within a single domain or a single partitioning algorithm cannot establish whether the same representation remains favorable for digits, characters, clothing, and natural objects. Similarly, retrieval-oriented embedding evaluations do not directly determine the class structure recovered by clustering.

Representation and algorithm choices for textual datasets have previously been examined~\cite{petukhova2025}. The present work extends this comparative perspective to images by evaluating nine representation pipelines across seven datasets and four clustering configurations, yielding 252 dataset--representation--clustering combinations. These pipelines include visual features extracted from vision--language checkpoints, vision-only references, and a caption-to-text route. While the latter builds on an established paradigm~\cite{tgic}, our contribution lies in systematically comparing these approaches over image domains and clustering algorithms. The unit of comparison is a specified extraction--clustering pipeline, rather than the best attainable clustering performance of each pretrained model.

We address four questions: (i) how do representations compare with the clustering algorithm held fixed; (ii) how do representation rankings vary with the clustering algorithm; (iii) how does caption-mediated representation compare with direct visual extraction in agreement with reference classes; and (iv) how sensitive are Ward agglomerative clustering-based comparisons to per-image L2 normalization? The 252 baseline dataset--representation--clustering combinations support a comparison of external agreement, internal cluster geometry, and recorded extraction time. The complementary normalization analysis comprises 63 paired comparisons at unchanged embedding dimensionality. The extraction components, pooling rules, and evaluation procedures define the specific pipelines under study. The comparison concerns fixed pretrained representations; it does not directly benchmark the trained clustering methods reviewed below.

\section{Background and Related Work}
\subsection{Visual and vision--language representations}
Vision--language representation learning can provide an image encoder whose output is useful independently of the text branch. CLIP learns an aligned image--text space through contrastive supervision~\cite{clip}. SigLIP2 extends image--text training with captioning, self-distillation, and masked-prediction objectives~\cite{siglip2}. For clustering, the relevant question is how the extracted image vectors organize the target collection. Their suitability for image--text matching does not necessarily translate into effective clustering under Euclidean distance, particularly when the feature extraction procedure differs from that used for image--text alignment.

Generative VLMs introduce further distinctions between the visual backbone, the interface connecting vision and language, and the language model. BLIP combines image--text understanding and generation objectives~\cite{blip}; BLIP-2 uses a querying transformer to connect pretrained visual and language components~\cite{blip2}. LLaVA-1.5 and Qwen3-VL support visual instruction following~\cite{llava,qwen}. These models can provide different image representations depending on the component and feature extraction procedure used. In particular, visual features can be extracted directly, without generating text. In our experiments, BLIP-2 produces 1,408-dimensional embeddings by averaging the visual backbone's hidden states, excluding the first token. Thus, the resulting vectors represent visual features, not language-model hidden states or generated captions. The specific components and feature extraction procedures used for each model are described in Section~\ref{sec:methods}.

Vision-only models offer an additional reference. DINOv3 learns visual features through self-supervision~\cite{dinov3}, whereas DiT targets document-image representation~\cite{dit}. Including them allows the comparison to encompass different pretraining objectives and domains. Since checkpoint architecture, training data, and extraction rules vary together, differences between these models and the VLM-derived features cannot be attributed solely to the presence of language supervision.

\subsection{VLMs trained as embedding models}
Extracting features from a pretrained generative vision--language model and explicitly training such a model to produce embeddings are two distinct approaches. Jiang et al. introduce VLM2Vec, a framework that uses contrastive learning to adapt vision--language models for embedding tasks, alongside the Massive Multimodal Embedding Benchmark (MMEB), which covers classification, visual question answering, multimodal retrieval, and visual grounding~\cite{vlm2vec}. VLM2Vec produces fixed-dimensional embeddings from images, text, or their combination, conditioned on task-specific instructions. This work highlights the distinction between using features extracted from a pretrained generative model and learning a representation space specifically optimized for embedding tasks.

Another approach is presented by mmE5, a framework for learning multilingual and multimodal embeddings that emphasizes the quality and diversity of training data~\cite{mme5}. It combines synthetic and labeled data, with a data synthesis procedure designed to improve task coverage, cross-modal alignment, and data fidelity. Together with VLM2Vec, mmE5 illustrates how dedicated embedding training differs from extracting representations from pretrained generative models. These approaches complement the direct visual feature extraction and caption-based embedding pipelines evaluated in our study, but involve additional training that falls outside its scope.

\subsection{Representation learning and adaptation for image clustering}

Image clustering methods can extend beyond applying a conventional clustering algorithm to fixed feature vectors by learning representations, adapting clustering assignments, or modifying feature construction. SCAN separates self-supervised representation learning from semantic clustering, using nearest-neighbor consistency and subsequent self-labeling~\cite{scan}. Although these stages are distinct, the clustering procedure itself involves learning rather than simply partitioning a fixed feature space.

Other approaches retain the pretrained visual encoder while introducing task-specific mechanisms. Li et al.~\cite{scp2025} propose SCP, a lightweight image clustering approach that trains a small clustering head on features extracted from augmented views of the same image, while keeping the pretrained visual encoder frozen and without requiring a text encoder. FUNGI instead augments pretrained embeddings with projected gradients of self-supervised objectives, without updating model weights~\cite{fungi2024}. Its feature components are individually L2-normalized before concatenation. These methods demonstrate that clustering can benefit from learned assignments or modified feature representations even when the visual backbone remains fixed. In contrast, our study applies conventional clustering algorithms to extracted representations, with a separate analysis of L2 normalization that does not introduce additional features or model training.

\subsection{Language-guided image clustering}

Language can also guide image clustering by providing semantic descriptions, specifying grouping criteria, or supporting representation adaptation. The \textit{Text-Guided Image Clustering} (TGIC) approach, proposed by Stephan et al.~\cite{tgic}, generates image captions or visual-question-answering outputs, embeds the resulting text, and compares clustering in visual and textual spaces. TGIC also uses prompts to incorporate domain knowledge and derives keyword-based cluster explanations. This approach provides a direct methodological precedent for our caption-mediated pipeline, which is evaluated alongside eight direct visual feature extraction pipelines.

In subsequent work, Stephan et al.~\cite{tgaicc} introduced TGAICC, a text-guided alternative clustering approach that addresses a different objective: discovering groupings that reflect user-defined interests rather than recovering a single reference partition. TGAICC generates prompt-dependent partitions and combines them through hierarchical and consensus clustering. This illustrates that agreement with predefined categories captures only one aspect of meaningful image clustering. Our evaluation focuses on agreement with the supplied dataset labels and does not examine alternative user-defined groupings.

More recent work couples language information with learned clustering methods. Li et al.~\cite{crossmodal2025} propose a self-enhanced image clustering framework (SEIC) based on cross-modal semantic consistency. SEIC first trains clustering heads by aligning image and text representations at multiple levels, then uses pseudo-labels to guide joint adaptation of the visual encoder and clustering heads. In contrast, Li et al.~\cite{gsec2026} introduce GSEC, which combines generative semantic guidance with a bi-layer ensemble designed to reduce bias and variance. Its inner ensemble integrates visual and textual information, while the outer layer aligns clustering predictions. While SEIC emphasizes cross-modal alignment followed by representation adaptation, GSEC combines generated semantic information with ensemble-based clustering. Both approaches extend beyond applying a single classical clustering algorithm to fixed pretrained features. Consequently, their reported improvements cannot be directly compared with our results without harmonizing representations, data splits, training procedures, and evaluation metrics.

Table~\ref{tab:related} summarizes these approaches in terms of representation construction, additional learning, and clustering strategy. The present study focuses on comparing fixed pretrained visual features and caption-mediated embeddings across datasets and conventional clustering algorithms, with a separate sensitivity analysis of L2 normalization. This design evaluates the behavior of specified representation--clustering pipelines without introducing learned clustering heads, encoder adaptation, or feature augmentation.

 % tables/related_work

\begin{landscape}
\begin{table}[tph!]
\centering
{\scriptsize{}%
\setlength{\tabcolsep}{6pt}
\renewcommand{\arraystretch}{1.15}
\caption{Selected related work compared in terms of representation or grouping strategy, additional training or processing, and relationship to the present study.}
\label{tab:related}
\begin{tabular}{@{}
    >{\raggedright\arraybackslash}p{.05\linewidth}
    >{\centering\arraybackslash}p{.02\linewidth}
    >{\centering\arraybackslash}p{.03\linewidth}
    >{\raggedright\arraybackslash}p{.28\linewidth}
    >{\raggedright\arraybackslash}p{.26\linewidth}
    >{\raggedright\arraybackslash}p{.26\linewidth}
    @{}}
\toprule
Study/ Model & Ref. & Year &
Representation or grouping strategy &
Additional training or processing &
Relation to this study \\
\midrule

VLM2Vec & \cite{vlm2vec} & 2025 &
Instruction-conditioned multimodal embeddings &
Contrastive embedding training &
Learns task-oriented embedding spaces \\

mmE5 & \cite{mme5} & 2025 &
Multimodal and multilingual embeddings &
Training on synthetic and labeled data &
Studies embedding training-data quality \\

SCAN & \cite{scan} & 2020 &
Self-supervised features and learned cluster assignments &
Representation learning, clustering, and self-labeling &
Reference for learned image clustering \\

SCP & \cite{scp2025}$^\star$ & 2025 &
Frozen visual features and positive pairs &
Small clustering-head training &
Fixed encoder with learned assignments \\

FUNGI & \cite{fungi2024} & 2024 &
Pretrained embeddings augmented with projected gradients &
Gradient-based feature construction; no weight updates &
Augments features; our L2 analysis preserves dimension \\

TGIC & \cite{tgic} & 2024 &
Generated captions/VQA outputs encoded as text &
Text gen. and encoding with pretrained models &
Direct antecedent of our caption route \\

TGAICC & \cite{tgaicc} & 2024 &
Prompt-conditioned partitions and consensus &
Hierarchical grouping and consensus aggregation &
Studies alternative grouping criteria \\

SEIC & \cite{crossmodal2025} & 2026 &
Cross-modal consistency and pseudo-labels &
Clustering-head training, then encoder adaptation &
Adapts representations to clustering \\

GSEC & \cite{gsec2026} & 2026 &
Generative semantic guidance and bi-layer ensemble &
Cross-modal ensemble integration and output alignment &
Combines semantic guidance with ensemble clustering \\

Present study & -- & -- &
Direct visual and caption-mediated embeddings &
No additional model training; conventional clustering &
Compares domains, clustering algorithms, and L2 normalization \\

\bottomrule
 \multicolumn{6}{l}{$^\star$Study published as a preprint.}

\end{tabular}}
\end{table}
\end{landscape}

\section{Methods}
\label{sec:methods}

\begin{figure}[!tbh]
\centering
\resizebox{\linewidth}{!}{%
\begin{tikzpicture}[
  box/.style={draw=black!65,rounded corners=2pt,align=center,font={\small\hyphenpenalty=10000\relax},inner sep=7pt},
  visual/.style={box,fill=blue!6,text width=5.99cm},
  captionroute/.style={box,fill=orange!9,text width=5.99cm},
  shared/.style={box,fill=black!3,text width=13.6cm},
  flow/.style={-{Stealth[length=2mm]},thick,draw=black!65}
  ]
  
\node[shared] (images) {\textbf{Test images from seven datasets}\\{\footnotesize{}MNIST, Fashion-MNIST, EMNIST, USPS, SVHN, CIFAR-10, CIFAR-100}};

\node[visual,below=7mm of images.south west,anchor=north west] (visual) {\textbf{Direct visual features (8)}\\
{\footnotesize{}CLIP, SigLIP2, BLIP, BLIP-2, LLaVA, Qwen3-VL, DINOv3, DiT}};

\node[captionroute,below=7mm of images.south east,anchor=north east] (caption) {\textbf{Caption + text (1)}\\
{\footnotesize{}\texttt{gpt-5.4-mini} (caption gen.) + 
\texttt{text-embedding-3-small} (text enc.)}};

\draw[flow] (images.south -| visual.north) -- (visual.north);
\draw[flow] (images.south -| caption.north) -- (caption.north);

\node[shared,below=7mm of $(visual.south)!0.5!(caption.south)$] (embeddings) 
{\textbf{Nine independent embedding spaces}\\
{\footnotesize{}Raw embeddings at original dimensionality}};

\draw[flow] (visual.south) -- (visual.south |- embeddings.north);
\draw[flow] (caption.south) -- (caption.south |- embeddings.north);

\node[shared,below=7mm of embeddings] (cluster) {\textbf{Four clustering configurations
}\\{\footnotesize{}K-means, K-means++, 
AHC-Ward, Fuzzy C-means
}};

\draw[flow] (embeddings) -- (cluster);

\node[shared,below=7mm of cluster] (evaluation) {\textbf{Evaluation: 252 combinations}\\{\footnotesize{}External and internal clustering metrics}};

\draw[flow] (cluster) -- (evaluation);

\end{tikzpicture}}

\caption{Experimental workflow across seven datasets, nine representation pipelines, and four clustering configurations ($7 \times 9 \times 4 = 252$ baseline combinations). The visual route includes six vision--language models and two vision-only models (DINOv3 and DiT), while the caption-mediated route generates and encodes image descriptions. Each representation is clustered independently, with the number of clusters $K$ determined by the reference class count. Reference labels are used only for evaluation and to determine $K$, not for feature extraction or fitting cluster assignments. The baseline uses raw, single-precision embeddings at their original dimensionality; a separate sensitivity analysis compares raw and L2-normalized embeddings with AHC-Ward clustering fixed.}
\label{fig:methodology}
\end{figure}
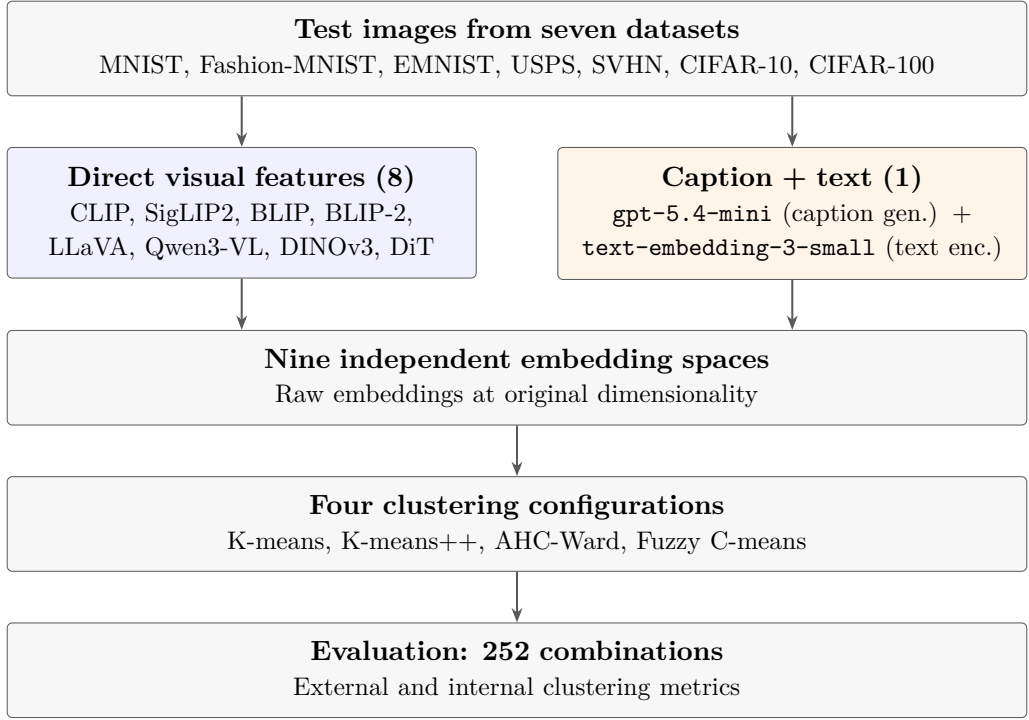

Figure~\ref{fig:methodology} summarizes the experimental workflow, comprising two representation routes followed by common clustering and evaluation stages. Each of the nine representations is evaluated independently with each clustering method, without combining embeddings from different models. The following subsections describe the datasets, the visual and caption-mediated feature extraction procedures, and the clustering and evaluation methods. They also document the consistency checks, the sensitivity analysis of L2 normalization, and the computational and visualization settings.

 % figures/methodology

\subsection{Datasets}

Seven image classification benchmarks are evaluated, spanning handwritten digits, characters, clothing, and natural images. MNIST~\cite{mnist} and USPS~\cite{usps} contain handwritten digits, while Fashion-MNIST~\cite{fashion} contains images of clothing items. EMNIST~\cite{emnist} extends handwritten character recognition to digits and letters; we use its balanced variant, comprising 47 classes. SVHN~\cite{svhn} contains digits cropped from street-view photographs, whereas CIFAR-10 and CIFAR-100~\cite{cifar} contain natural images organized into 10 and 100 classes, respectively.

All experiments use the complete test partition of each dataset, without subsampling. Training partitions are not used for feature extraction or clustering. Table~\ref{tab:datasets} reports the number of evaluated images and reference classes for each dataset. These benchmarks provide a range of image domains and category structures for assessing the clustering behavior of pretrained representations.

 % tables/datasets-ng
\begin{table}[tb]
\centering
{\scriptsize{}%

\caption{Evaluated test partitions of the seven image benchmarks, showing the number of images and reference classes.}
\label{tab:datasets}
\begin{tabular}{llrr}
\toprule
Dataset & Ref. & Images & Classes \\
\midrule
MNIST & \cite{mnist} & 10,000 & 10 \\
USPS & \cite{usps} & 2,007 & 10 \\
Fashion-MNIST & \cite{fashion} & 10,000 & 10 \\
EMNIST (balanced) & \cite{emnist} & 18,800 & 47 \\
SVHN & \cite{svhn} & 26,032 & 10 \\
CIFAR-10 & \cite{cifar} &  10,000 & 10 \\
CIFAR-100 & \cite{cifar} & 10,000 & 100 \\
\bottomrule
\end{tabular}}
\end{table}

\subsection{Image preprocessing and feature extraction}

Table~\ref{tab:checkpoints} lists the pretrained checkpoints used for feature extraction. Grayscale images are converted to three channels by replication, while RGB images retain their original channels. Each image is then processed using the preprocessing pipeline associated with the corresponding model. All models are used without dataset-specific fine-tuning, allowing their pretrained representations to be evaluated directly.

 % tables/checkpoints
\begin{table}[tb]
\centering
{\scriptsize{}%

\caption{Pretrained checkpoint identifiers. The extracted components and embedding dimensions are specified in Table~\ref{tab:extraction-protocol}.}\label{tab:checkpoints}
\begin{tabular}{ll}
\toprule
Representation & Checkpoint identifier \\
\midrule
CLIP & \texttt{openai/clip-vit-base-patch32} \\
SigLIP2 & \texttt{google/siglip2-base-patch16-224} \\
BLIP & \texttt{Salesforce/blip-image-captioning-base} \\
BLIP-2 & \texttt{Salesforce/blip2-opt-2.7b} \\
LLaVA & \texttt{llava-hf/llava-1.5-7b-hf} \\
Qwen3-VL & \texttt{Qwen/Qwen3-VL-8B-Instruct} \\
DINOv3 & \texttt{facebook/dinov3-vits16-pretrain-lvd1689m} \\
DiT & \texttt{microsoft/dit-base} \\
Caption + text & \texttt{gpt-5.4-mini} + \texttt{text-embedding-3-small} \\
\bottomrule
\end{tabular}}
\end{table}

For most locally executed models, image embeddings are obtained by averaging selected visual token representations. The extraction procedure varies according to the model architecture and the interface through which its features are accessed. For CLIP, patch-level hidden states are averaged, followed by the vision post-layer normalization and projection to a 512-dimensional embedding space. 
SigLIP2 uses the mean of all visual patch hidden states. BLIP, BLIP-2, and DiT use the mean of the visual hidden states after excluding the first token.
DINOv3 follows a similar procedure, additionally excluding the configured register tokens before averaging.

For LLaVA, the representation is obtained by averaging the projected visual patch tokens after excluding the CLS token retained by the full feature-selection strategy. For Qwen3-VL, all tokens in the returned image-feature sequence are averaged, without excluding the first entry.

The token-averaging procedure shared by these extraction methods can be expressed as

\begin{equation}
 z_m(x)=\frac{1}{|\mathcal{T}_m(x)|}\sum_{t\in\mathcal{T}_m(x)} h_{m,t}(x),
\end{equation}

\noindent where $h_{m,t}(x)$ denotes the hidden representation of token $t$ for image $x$ under model $m$, and $\mathcal{T}_m(x)$ is the set of tokens retained for aggregation. The resulting vector $z_m(x)$ is used as the image representation, with any additional model-specific transformations applied as described above. 

The same extraction procedure is applied consistently to all images within each model. The extracted components, aggregation rules, and resulting embedding dimensions are summarized in Table~\ref{tab:extraction-protocol}, while implementation details and version-dependent behavior are documented in the supplementary protocol.

 % tables/extraction_protocol
\begin{table}[tb]
\centering
{\scriptsize{}%

\caption{Representation components and aggregation used in the main comparison. Dimensions refer to the vectors supplied to clustering. The baseline uses no additional vector normalization; the sensitivity analysis applies per-image L2 normalization. The routes describe the evaluated implementations, not an optimization over alternative outputs of each model.}
\label{tab:extraction-protocol}
\begin{tabular}{lp{9.5cm}r}
\toprule
Representation & Component and aggregation & Dim. \\
\midrule
CLIP & Visual patch-state mean, followed by vision post-layer normalization and visual projection & 512 \\
SigLIP2 & Mean of all visual patch states 
& 768 \\
BLIP & Visual-state mean after the first token & 768 \\
BLIP-2 & Visual-backbone state mean after the first token, rather than language-model states & 1,408 \\
LLaVA & Mean after the first entry of the returned image-feature sequence & 4,096 \\
Qwen3-VL & Mean of all tokens in each returned image-feature sequence
& 4,096 \\
DINOv3 & Visual-state mean excluding the initial token and register tokens & 384 \\
DiT & Visual-state mean after the first token & 768 \\
Caption + text & Generated image description encoded by a text-embedding model & 1,536 \\
\bottomrule
\end{tabular}}
\end{table}

\subsection{Caption-mediated representation}

The ninth pipeline uses \texttt{gpt-5.4-mini} to generate one short description per image, followed by \texttt{text-embedding-3-small} to obtain a 1,536-dimensional text vector. This pipeline is denoted \captionmodel{} in Tables~\ref{tab:checkpoints}~and~\ref{tab:extraction-protocol} and for the remainder of this paper to distinguish its output from direct visual embeddings.

Caption generation uses dataset-dependent instructions. Digit prompts request identity and stroke properties; the clothing prompt requests category, silhouette, and texture; natural-image prompts request the main object or scene class and context. No ground-truth label is inserted into the caption request, but these instructions provide task-specific semantic guidance that the direct visual pipelines do not receive. Consequently, the comparison evaluates complete pipelines rather than isolating the effect of representation architecture. Extraction timings include caption generation for every image, without reuse of previously generated descriptions. Caption recognition accuracy was not evaluated separately. Generation used the API defaults without explicit temperature or seed control; deterministic caption generation is therefore not assumed.

\subsection{Clustering algorithms and configurations}

We evaluate three clustering methods: K-means, fuzzy c-means, and agglomerative hierarchical clustering (AHC) with Ward linkage. K-means is considered under two initialization strategies, yielding four clustering configurations in total. All methods operate on the extracted image embeddings without additional representation learning.

K-means partitions observations into $K$ clusters by iteratively assigning each observation to its nearest centroid and updating the centroids to minimize within-cluster squared Euclidean distances~\cite{kmeans}. We consider random initialization and K-means++ initialization, which selects initial centroids using distance-weighted sampling to favor well-separated starting points~\cite{kmeanspp}. Both configurations optimize the same objective but may produce different partitions because of their initialization and restart settings.

Fuzzy c-means extends centroid-based clustering by assigning each observation a degree of membership in every cluster instead of requiring an exclusive assignment during optimization~\cite{fcm1984}. Cluster prototypes and membership degrees are updated iteratively. For evaluation against reference labels, the resulting memberships are converted into a hard partition by assigning each observation to the cluster with its highest membership.

AHC constructs a hierarchy by successively merging clusters. We use Ward linkage, which selects at each step the pair of clusters whose merger produces the smallest increase in total within-cluster sum of squares~\cite{ward1963}. The resulting hierarchy is cut at the prespecified number of clusters to obtain a partition.

For all configurations, the number of clusters $K$ is set to the number of distinct reference classes in each dataset. This corresponds to an oracle-$K$ evaluation, i.e., the reference labels determine the cluster count but are not used to fit cluster assignments. The baseline uses single-precision (32-bit) floating-point embeddings at their original dimensionality, without additional feature standardization, vector normalization, or dimensionality reduction. A separate sensitivity analysis examines the effect of L2 normalization with AHC-Ward clustering fixed.

All four clustering configurations use Euclidean distance. Both K-means configurations use random seed 0. K-means++ uses one initialization, while random-initialized K-means uses ten, retaining the solution with the lowest inertia. Since the K-means configurations differ in both initialization strategy and number of initializations, their performance differences cannot be attributed solely to initialization strategy. Fuzzy c-means uses the library defaults for its remaining parameters, including initialization and convergence criteria. Each stochastic clustering configuration is evaluated once, without estimating run-to-run variability. The software environment and implementation-dependent settings are documented in Section~\ref{sec:methods:computational}.

\subsection{Evaluation and aggregation}

Clustering quality is assessed using external and internal evaluation criteria. External metrics measure agreement between the inferred clusters and the reference class labels, while internal metrics assess cluster cohesion and separation without relying on those labels. Since the objective of this study is to recover the reference classes, external evaluation is prioritized.

The primary evaluation metric throughout this study is the adjusted Rand index (ARI)~\cite{hubert1985}, which measures agreement between two partitions based on whether pairs of observations are assigned to the same or different clusters. ARI adjusts for agreement expected by chance, attaining a maximum of 1 for identical partitions, values near 0 for chance-level agreement, and potentially negative values for agreement below chance. As ARI is invariant to cluster label permutations, it does not require matching inferred clusters to reference classes.

Three additional metrics provide complementary assessments. V-measure~\cite{vmeasure2007} is an external metric defined as the harmonic mean of homogeneity and completeness, assessing whether clusters contain observations from a single reference class and whether observations from the same class are grouped together, respectively. Class-frequency-weighted F1 is also computed after greedily matching inferred clusters to reference classes according to the squared Euclidean distances between their centroids in the corresponding embedding space. Since this matching does not necessarily maximize class agreement, the metric is denoted \emph{F1 (greedy)}. The silhouette coefficient~\cite{silhouette1987} provides an internal assessment of cluster cohesion and separation in the original embedding space. Because different representations define distinct feature spaces, silhouette scores are not used to rank representations. These complementary metrics are reported primarily in the appendices and discussed in the main text where relevant.

The main representation comparison holds AHC-Ward clustering fixed and ranks representations by ARI within each dataset, assigning rank 1 to the highest ARI and average ranks to ties. Mean ranks, mean and median ARI, paired ARI differences, and the number of datasets on which each representation achieves the highest ARI are reported. To assess the effect of clustering configuration, ARI is also averaged separately for each of the four configurations and across all configurations. Dataset-level averages assign equal weight to each of the seven datasets, while averages across configurations additionally assign equal weight to each configuration. Finally, the highest ARI observed among the 36 representation--clustering configurations is reported for each dataset, alongside the corresponding V-measure and F1 (greedy). These maxima are selected retrospectively and represent the best observed combinations rather than unbiased estimates of model-selection performance. All aggregate comparisons are descriptive. 

\subsection{Sensitivity to L2 normalization}
\label{sec:methods:l2}

For each of the 63 dataset--representation pairs, we compare the original embedding $z_i$ with
\begin{equation}
    \widetilde{z}_i=\frac{z_i}{\lVert z_i\rVert_2}.
\end{equation}
This transformation gives each image vector unit Euclidean norm while preserving its direction and original dimensionality. It is distinct from PCA, feature-wise standardization, and the internal layer normalization used by some encoders. All input vectors have nonzero norm. Norms and division are computed in double precision, followed by conversion to single precision for clustering.

AHC-Ward uses the known class count, a complete hierarchy, and no connectivity constraint, with identical settings for raw and normalized vectors. For unit vectors, squared Euclidean distance is $\lVert\widetilde{z}_i-\widetilde{z}_j\rVert_2^2=2(1-\widetilde{z}_i^\top\widetilde{z}_j)$; AHC-Ward nevertheless retains its Euclidean within-cluster variance objective. The comparison consists of 126 additional fits: 63 raw-vector controls and 63 normalized-vector conditions, separate from the original 252-combination benchmark. No feature extraction or caption generation is repeated.

Clustering performance is primarily compared using ARI for both the original and L2-normalized embeddings. For each dataset--representation pair, the effect of normalization is quantified as the difference between the scores obtained with L2-normalized embeddings and those obtained from the corresponding raw-vector controls, recomputed under identical conditions. Mean differences assign equal weight to each of the seven datasets.

\subsection{Visualization and recorded extraction time}

For visualization, we apply PCA retaining 99\% of variance before two-dimensional t-SNE, using perplexity 30, learning rate 150, angle 0.2, and random seed 42. Displayed t-SNE axes are independently rescaled to $[0,1]$. The resulting projections illustrate the structure of the embedding spaces and are not used for clustering or evaluation.

Recorded local extraction times begin after model loading and include the image-processing and feature-extraction loop. For the caption-mediated pipeline, timings include caption generation and text-embedding requests, as well as network latency. The reported values represent elapsed extraction time per image under the respective local and remote execution conditions.

\subsection{Consistency and reproducibility checks}

For the baseline experiments, numerical checks covered all 781,551 embeddings, corresponding to nine representations of 86,839 test images. Sample counts and feature dimensions matched the experimental design, all vectors had finite components and nonzero norm, and reference-label distributions agreed across representations within each dataset. All 252 baseline dataset--representation--clustering combinations also produced complete sets of finite evaluation metrics.

For the L2 normalization analysis, all 63 original AHC-Ward configurations were recomputed in the same environment as their normalized counterparts. Both ARI and V-measure matched the original baseline values within $10^{-12}$, with a maximum absolute difference of $2.22\times10^{-16}$. Evaluation metrics for all 126 additional partitions were independently recalculated from their predicted cluster assignments.

\subsection{Computational platform}
\label{sec:methods:computational}

The local experiments were conducted on an NVIDIA DGX Spark Founders Edition with 4~TB NVMe storage. The system uses the NVIDIA GB10 Grace Blackwell Superchip, integrating a Blackwell GPU and a 20-core Arm CPU (10 Cortex-X925 and 10 Cortex-A725 cores). It provides 128~GB of LPDDR5x unified memory shared by the CPU and GPU, with a specified memory bandwidth of 273~GB/s~\cite{dgxspark}. The OpenAI caption and text-embedding models were accessed through a remote API, while the remaining feature extraction, clustering, and evaluation procedures were executed locally.

The software environment used Python 3.12.3~\cite{python3}, NumPy 2.5.1~\cite{numpy}, SciPy 1.18.0~\cite{scipy}, scikit-learn 1.9.0~\cite{scikit}, and fuzzy-c-means 1.7.2~\cite{fuzzylib} for numerical processing and clustering. Local feature extraction used PyTorch 2.13.0~\cite{pytorch}, torchvision 0.28.0~ \cite{torchvision2016}, and Transformers 5.14.1~\cite{transformers}. Matplotlib 3.10.9~\cite{matplotlib} was used for visualization, and the OpenAI API was accessed through the openai Python library version 2.45.0~\cite{openaipython}. 
The same library versions were used throughout the baseline and sensitivity experiments.

\section{Results and Discussion}
\label{sec:results}

This section presents and discusses the results of the experimental evaluation described in Section~\ref{sec:methods}. We begin by comparing the nine representation pipelines in the seven datasets with fixed AHC-Ward clustering, followed by an analysis of representation rankings and dataset-level performance (Section~\ref{sec:results:fixed}). We then examine the sensitivity of clustering quality to L2 normalization of the embeddings (Section~\ref{sec:results:l2}) and assess how performance varies between the four clustering configurations (Section~\ref{sec:results:cluconf}). Next, we compare representations obtained directly from visual features with those derived from generated image descriptions (Section~\ref{sec:results:viscap}). We then consider embedding dimensionality, recorded extraction times, and illustrative visualizations of the resulting feature spaces (Section~\ref{sec:results:dimtimevis}). Finally, we synthesize the main findings and their implications for selecting representation--clustering pipelines (Section~\ref{sec:results:synthesis}).

\subsection{Representation performance with fixed clustering}
\label{sec:results:fixed}

Tables~\ref{tab:ward-ari} presents the clustering performance of the nine representation pipelines on the seven datasets, measured using ARI. AHC-Ward clustering is held fixed, allowing the comparison to focus on differences between representations. 

 % tables/ward_ari
\begin{table}[!b]
\centering
{\scriptsize{}%
\caption{Adjusted Rand index (ARI) with a fixed AHC-Ward clustering algorithm. Bold values indicate the best representation for each dataset.}
\label{tab:ward-ari}
\begin{tabular}{lrrrrrrr}
\toprule
Representation & MNIST & Fashion & EMNIST & USPS & SVHN & CIFAR-10 & CIFAR-100 \\
\midrule
CLIP & 0.511 & 0.534 & 0.283 & 0.426 & 0.013 & 0.508 & 0.186 \\
SigLIP2 & 0.479 & 0.529 & 0.332 & 0.687 & 0.012 & 0.575 & 0.311 \\
BLIP & 0.614 & 0.515 & 0.324 & 0.579 & 0.010 & 0.453 & 0.216 \\
BLIP-2 & 0.588 & 0.526 & 0.327 & 0.685 & 0.033 & \textbf{0.925} & \textbf{0.591} \\
LLaVA & 0.176 & 0.237 & 0.106 & 0.266 & 0.005 & 0.340 & 0.168 \\
Qwen3-VL & \textbf{0.925} & \textbf{0.573} & \textbf{0.434} & \textbf{0.717} & \textbf{0.097} & 0.777 & 0.475 \\
DINOv3 & 0.468 & 0.485 & 0.303 & 0.612 & 0.010 & 0.655 & 0.287 \\
DiT & 0.208 & 0.238 & 0.083 & 0.153 & 0.003 & 0.025 & 0.012 \\
Caption + text & 0.793 & 0.193 & 0.104 & 0.694 & 0.015 & 0.273 & 0.111 \\
\bottomrule
\end{tabular}}
\end{table}

The ARI results reveal a domain-dependent pattern, with the relative performance of the representations varying according to the characteristics of the datasets. Qwen3-VL achieves the highest ARI on five datasets (MNIST, Fashion-MNIST, EMNIST, USPS, and SVHN), whereas BLIP-2 leads on the two natural-image datasets (CIFAR-10 and CIFAR-100). This suggests that no single representation is consistently superior in all the tested domains, although these two models stand out among the evaluated pipelines.

On MNIST, Qwen3-VL achieves an ARI of 0.925, followed by the caption-mediated pipeline at 0.793 and BLIP at 0.614. However, its performance is substantially lower on EMNIST, where the highest ARI is 0.434. This contrast illustrates the greater difficulty of recovering the reference classes in the more diverse character dataset compared with the ten handwritten digit classes of MNIST.

For natural images, BLIP-2 achieves an ARI of 0.925 on CIFAR-10, outperforming Qwen3-VL (0.777) and DINOv3 (0.655). BLIP-2 and Qwen3-VL also occupy the leading positions on CIFAR-100, although with lower ARI values, reflecting the greater difficulty of recovering its 100 reference classes. These results highlight the effectiveness of the visual features extracted from BLIP-2 for the evaluated natural-image datasets, despite the model being used solely as a visual feature extractor rather than through its language-generation capabilities.

SVHN presents a particularly challenging case. Even Qwen3-VL, the best-performing representation on this dataset, achieves an ARI of only 0.097, with the remaining pipelines performing substantially worse. This contrasts sharply with the results on MNIST and USPS, despite all three datasets involving digits. Unlike the isolated handwritten digits in MNIST and USPS, SVHN contains digits extracted from street-view photographs, introducing greater variability in appearance, backgrounds, and surrounding visual content. The results suggest that these differences pose substantial challenges to recovering digit classes from the evaluated representations.

The V-measure results in appendix Table~\ref{tab:ward-vmeasure} provide a complementary assessment of agreement with the reference classes and generally corroborate the ARI findings: Qwen3-VL and BLIP-2 remain the leading representations, with the same best-performing model identified by both metrics on six of the seven datasets. The exception is Fashion-MNIST, where BLIP-2 achieves the highest V-measure (0.720), despite Qwen3-VL obtaining the highest ARI (0.573). Overall, the agreement between the two evaluation metrics reinforces the observed domain-dependent differences in clustering performance.

Table~\ref{tab:ward-ranks} summarizes the relative performance of the nine representations with AHC-Ward clustering, including their mean ranks, mean and median ARI, and number of dataset wins. Qwen3-VL achieves the best mean rank (1.286), followed by BLIP-2 (2.714) and SigLIP2 (3.714). Qwen3-VL outperforms BLIP-2 on five of the seven datasets, with a mean paired ARI difference of 0.046 and a median paired difference of 0.047. However, BLIP-2 performs better on both CIFAR datasets, highlighting the domain-dependent nature of their relative performance. The mean and median ARI values provide descriptive summaries over the evaluated benchmarks, but should be interpreted with caution, as the datasets differ substantially in their characteristics and clustering difficulty.

 % tables/ward_rankings
\begin{table}[!t]
\centering
{\scriptsize{}%
\caption{Descriptive summary of representation performance with AHC-Ward clustering held fixed. Ranks are computed within each dataset (1 indicates the best performance) and averaged equally across the seven datasets. Wins indicate the number of datasets on which each representation achieves the highest ARI. Mean and median ARI are reported as descriptive summaries of the evaluated benchmarks and should be interpreted in light of differences in dataset characteristics and clustering difficulty.}
\label{tab:ward-ranks}
\begin{tabular}{lrrrr}
\toprule
Representation & Mean rank & Mean ARI & Median ARI & Wins \\
\midrule
Qwen3-VL & 1.286 & 0.571 & 0.573 & 5 \\
BLIP-2 & 2.714 & 0.525 & 0.588 & 2 \\
SigLIP2 & 3.714 & 0.418 & 0.479 & 0 \\
BLIP & 5.000 & 0.387 & 0.453 & 0 \\
CLIP & 5.000 & 0.352 & 0.426 & 0 \\
DINOv3 & 5.286 & 0.403 & 0.468 & 0 \\
Caption + text & 5.714 & 0.312 & 0.193 & 0 \\
LLaVA & 7.714 & 0.185 & 0.176 & 0 \\
DiT & 8.571 & 0.103 & 0.083 & 0 \\
\bottomrule
\end{tabular}}
\end{table}

\subsection{Effect of L2 normalization} 
\label{sec:results:l2}

Table~\ref{tab:l2-ari} reports ARI after L2 normalization, and Table~\ref{tab:l2-summary} summarizes paired changes relative to the raw-vector controls. Normalization increases ARI in 35 of the 63 pairs, decreases it in 26, and leaves two unchanged within numerical tolerance. Thus, its effect is representation- and dataset-dependent. The highest-ARI representation remains unchanged in every dataset: Qwen3-VL leads on five datasets and BLIP-2 on the two CIFAR datasets. This stability concerns the identity of the leading representation, not the full ordering or the magnitude of its advantage.

 % tables/l2_ari
\begin{table}[!t]
\centering
{\scriptsize{}%
\caption{ARI with AHC-Ward after per-image L2 normalization, retaining the original embedding dimensions. Bold marks the highest value in each dataset before rounding.}
\label{tab:l2-ari}
\begin{tabular}{lrrrrrrr}
\toprule
Representation & MNIST & Fashion & EMNIST & USPS & SVHN & CIFAR-10 & CIFAR-100 \\
\midrule
CLIP & 0.474 & 0.517 & 0.284 & 0.472 & 0.011 & 0.411 & 0.189 \\
SigLIP2 & 0.534 & 0.508 & 0.339 & 0.675 & 0.023 & 0.611 & 0.310 \\
BLIP & 0.670 & 0.488 & 0.301 & 0.555 & 0.014 & 0.486 & 0.213 \\
BLIP-2 & 0.507 & 0.564 & 0.333 & 0.716 & 0.035 & \textbf{0.911} & \textbf{0.584} \\
LLaVA & 0.312 & 0.451 & 0.159 & 0.372 & 0.010 & 0.391 & 0.200 \\
Qwen3-VL & \textbf{0.933} & \textbf{0.583} & \textbf{0.440} & \textbf{0.841} & \textbf{0.098} & 0.909 & 0.516 \\
DINOv3 & 0.502 & 0.565 & 0.308 & 0.539 & 0.006 & 0.640 & 0.297 \\
DiT & 0.161 & 0.217 & 0.092 & 0.172 & 0.000 & 0.024 & 0.011 \\
Caption + text & 0.793 & 0.193 & 0.103 & 0.694 & 0.014 & 0.272 & 0.110 \\
\bottomrule
\end{tabular}}
\end{table}

 % tables/l2_summary
\begin{table}[!t]
\centering
{\scriptsize{}%
\caption{Mean ARI with AHC-Ward, equally weighted across seven datasets. $\Delta$ is L2 minus raw. Higher/lower/equal counts compare L2 with raw within the same representation and dataset, using a numerical tolerance of $10^{-12}$.} 
\label{tab:l2-summary}
\begin{tabular}{lrrrrrr}
\toprule
Representation & Raw ARI & L2 ARI & $\Delta$ & Higher & Lower & Equal \\
\midrule
CLIP & 0.352 & 0.337 & -0.0147 & 3 & 4 & 0 \\
SigLIP2 & 0.418 & 0.428 & +0.0104 & 4 & 3 & 0 \\
BLIP & 0.387 & 0.390 & +0.0022 & 3 & 4 & 0 \\
BLIP-2 & 0.525 & 0.521 & -0.0035 & 4 & 3 & 0 \\
LLaVA & 0.185 & 0.271 & +0.0850 & 7 & 0 & 0 \\
Qwen3-VL & 0.571 & 0.617 & +0.0460 & 7 & 0 & 0 \\
DINOv3 & 0.403 & 0.408 & +0.0054 & 4 & 3 & 0 \\
DiT & 0.103 & 0.097 & -0.0064 & 2 & 5 & 0 \\
Caption + text & 0.312 & 0.312 & -0.0004 & 1 & 4 & 2 \\
\bottomrule
\end{tabular}}
\end{table}

LLaVA benefits on all seven datasets, with mean ARI increasing from 0.185 to 0.271. Its largest gain is on Fashion-MNIST ($\Delta\mathrm{ARI}=+0.213$). Qwen3-VL likewise improves on all seven datasets, with its mean ARI increasing from 0.571 to 0.617; on CIFAR-10, its ARI rises from 0.777 to 0.909, approaching BLIP-2's normalized ARI of 0.911. Conversely, CLIP's mean ARI falls from 0.352 to 0.337, with a decrease of 0.097 on CIFAR-10. BLIP-2 has a small mean decrease, from 0.525 to 0.521, despite improving on four datasets. These examples show why neither universal benefit nor universal harm should be attributed to normalization.

SVHN remains challenging: the highest normalized ARI is 0.098, obtained by Qwen3-VL, compared with 0.097 for the raw vectors. The principal domain-dependent pattern thus persists under this change in geometry. However, changes such as LLaVA's gains and the reduced CIFAR-10 gap demonstrate that quantitative comparisons should state whether vector norms are retained.

Complete normalized V-measure results are provided in appendix Table~\ref{tab:l2-v_measure}. After L2 normalization, the highest-V-measure representations match the ARI winners on all seven datasets, resolving the previous discrepancy on Fashion-MNIST.

\subsection{Effect of clustering configuration}
\label{sec:results:cluconf}

The aggregate comparison in Table~\ref{tab:means} shows that AHC-Ward achieves the highest mean ARI for eight of the nine representations. The sole exception is Qwen3-VL, for which random-initialized K-means increases mean ARI from 0.571 to 0.588, further strengthening its lead over the other representations. More generally, random-initialized K-means achieves a higher mean ARI than K-means++ for seven of the nine representations, with CLIP favoring K-means++ and LLaVA showing equal values after rounding. This pattern may partly reflect the different restart budgets: random-initialized K-means uses ten initializations, retaining the solution with the lowest inertia, whereas K-means++ uses only one. Therefore, the observed differences characterize the evaluated configurations rather than the relative merits of their initialization strategies alone.

 % tables/algorithm_means
\begin{table}[!t]
\centering
{\scriptsize{}%
\caption{ARI averaged equally across seven datasets, separately for each clustering configuration. The final column also averages equally across configurations; bold indicates the highest mean ARI for each representation among the four clustering configurations, selected before rounding. These averages summarize performance across datasets with differing characteristics and clustering difficulty.}
\label{tab:means}
\begin{tabular}{lrrrrr}
\toprule
Representation & K-means++ & K-means (random) & AHC-Ward & Fuzzy c-means & Mean \\
\midrule
CLIP & 0.314 & 0.308 & \textbf{0.352} & 0.062 & 0.259 \\
SigLIP2 & 0.386 & 0.400 & \textbf{0.418} & 0.084 & 0.322 \\
BLIP & 0.378 & 0.385 & \textbf{0.387} & 0.080 & 0.308 \\
BLIP-2 & 0.474 & 0.506 & \textbf{0.525} & 0.081 & 0.397 \\
LLaVA & 0.170 & 0.170 & \textbf{0.185} & 0.053 & 0.145 \\
Qwen3-VL & 0.546 & \textbf{0.588} & 0.571 & 0.085 & 0.447 \\
DINOv3 & 0.376 & 0.377 & \textbf{0.403} & 0.105 & 0.315 \\
DiT & 0.094 & 0.095 & \textbf{0.103} & 0.086 & 0.094 \\
Caption + text & 0.292 & 0.298 & \textbf{0.312} & 0.056 & 0.239 \\
\bottomrule
\end{tabular}}
\end{table}

Fuzzy c-means performs substantially worse than the other configurations, with mean ARIs ranging from 0.053 to 0.105. DINOv3 achieves the highest fuzzy c-means mean (0.105), despite Qwen3-VL leading under the other three configurations. Notably, DINOv3 also produces the lowest-dimensional embeddings (384 dimensions), suggesting that dimensionality may contribute to the poor fuzzy c-means results. This is consistent with known difficulties of fuzzy c-means in high-dimensional spaces, where cluster prototypes can converge toward the overall data centroid~\cite{fuzzyhighdim}. An exploratory diagnostic provides further support: on MNIST, reducing Qwen3-VL embeddings from 4,096 to five dimensions using PCA (retaining 49.8\% of explained variance) increased fuzzy c-means ARI from 0.092 to 0.627 under the same library defaults. The diagnostic suggests that dimensionality reduction can substantially improve fuzzy c-means performance in this setting, although it does not isolate dimensionality from changes in feature geometry or initialization sensitivity. Note that PCA was deliberately excluded from the main clustering experiments to preserve the original dimensionality of each extracted representation and avoid introducing an additional preprocessing variable, keeping the evaluation focused on the specified representations, datasets, and clustering configurations. 

Table~\ref{tab:best} provides a complementary dataset-level perspective, identifying the highest observed ARI among the 36 representation--configuration combinations evaluated for each dataset. Qwen3-VL achieves the highest ARI on five datasets, while BLIP-2 leads on both CIFAR datasets, preserving the representation-level pattern observed with fixed AHC-Ward clustering. However, the clustering configuration associated with the highest ARI varies: AHC-Ward is selected on three datasets, random-initialized K-means on three, and K-means++ on one. The highest ARI overall is 0.970, obtained by BLIP-2 with random-initialized K-means on CIFAR-10, compared with 0.925 under AHC-Ward. Random-initialized K-means also produces the highest ARIs for Fashion-MNIST (0.630) and USPS (0.759), both with Qwen3-VL, while K-means++ achieves the highest ARI on MNIST (0.941), also with Qwen3-VL. Thus, although AHC-Ward leads most representation-level averages, other configurations achieve higher agreement on individual datasets. These maxima are identified retrospectively using reference labels and describe the best observed combinations within the evaluated grid, rather than the performance of a configuration selected without access to those labels.

 % tables/best
\begin{table}[!t]
\centering
{\scriptsize{}%
\caption{Observed maximum-ARI configuration per dataset, selected retrospectively among 36 configurations. V-measure and weighted F1 refer to that same configuration.}
\label{tab:best}
\begin{tabular}{lllrrr}
\toprule
Dataset & Representation & Clustering config. & ARI & V-measure & F1 (greedy) \\
\midrule
MNIST & Qwen3-VL & K-means++ & 0.941 & 0.933 & 0.973 \\
Fashion-MNIST & Qwen3-VL & K-means (random) & 0.630 & 0.728 & 0.751 \\
EMNIST & Qwen3-VL & AHC-Ward & 0.434 & 0.687 & 0.554 \\
USPS & Qwen3-VL & K-means (random) & 0.759 & 0.834 & 0.813 \\
SVHN & Qwen3-VL & AHC-Ward & 0.097 & 0.158 & 0.217 \\
CIFAR-10 & BLIP-2 & K-means (random) & 0.970 & 0.964 & 0.986 \\
CIFAR-100 & BLIP-2 & AHC-Ward & 0.591 & 0.822 & 0.739 \\
\bottomrule
\end{tabular}}
\end{table}

\subsection{Direct visual versus caption-mediated representations}
\label{sec:results:viscap}

The caption-mediated pipeline achieves relatively high agreement with reference classes on MNIST and USPS, with AHC-Ward ARIs of 0.793 and 0.694, respectively (Table~\ref{tab:ward-ari}). However, its performance is substantially lower on Fashion-MNIST (0.193), EMNIST (0.104), CIFAR-10 (0.273), and CIFAR-100 (0.111). Per-image L2 normalization has little effect on these results, with mean ARI remaining approximately 0.312 over the seven datasets (Table~\ref{tab:l2-summary}). Thus, explicitly requesting object or character identity does not consistently produce text embeddings that recover the reference classes.

Unlike direct visual feature extraction, the caption-mediated approach introduces an intermediate step in which image content is selected and verbalized before being encoded as text. This semantic bottleneck may preserve class-relevant information for isolated handwritten digits while losing distinctions important in other domains. Possible explanations for the weaker results include errors in recognizing depicted objects or characters, variation in caption wording, and descriptions emphasizing attributes shared across reference classes. Identifying their relative contributions would require a separate analysis of caption accuracy and the class-discriminative information retained during verbalization, beyond the present study's comparison of complete representation--clustering pipelines. A useful extension would compare common and dataset-specific caption prompts, as well as different caption-generation models paired with the same text encoder.

\subsection{Embedding dimensionality, extraction time, and visualization}
\label{sec:results:dimtimevis}

Table~\ref{tab:cost} reports pooled embedding extraction times, alongside the embedding dimensions previously specified in Table~\ref{tab:extraction-protocol} for ease of comparison. Qwen3-VL and LLaVA both produce 4,096-dimensional embeddings, yet their mean ARIs across the complete evaluation grid differ substantially (0.447 and 0.145, respectively). Conversely, DINOv3 uses only 384 dimensions and achieves a mean ARI of 0.315, providing a comparatively compact representation with higher agreement than several larger embeddings. These observations suggest that dimensionality alone does not explain the differences in clustering performance across representations. As discussed in Section~\ref{sec:results:cluconf}, dimensionality reduction can influence clustering outcomes, but systematically evaluating this effect would require additional experiments beyond the present comparison of representations at their original dimensionality.

 % tables/cost
\begin{table}[!b]
\centering
{\scriptsize{}%

\caption{Pooled embedding extraction time (total recorded seconds divided by total images), with embedding dimensions repeated from Table~\ref{tab:extraction-protocol} for ease of analysis. Local timings exclude model loading, while API timings include caption generation and network latency. Local extraction was performed on an NVIDIA DGX Spark; timings reflect the respective local and remote execution conditions.}

\label{tab:cost}
\begin{tabular}{lrr}
\toprule
Representation & Dimension & Recorded ms/image \\
\midrule
CLIP & 512 & 1.49 \\
SigLIP2 & 768 & 4.18 \\
BLIP & 768 & 17.91 \\
BLIP-2 & 1,408 & 66.07 \\
LLaVA & 4,096 & 10.63 \\
Qwen3-VL & 4,096 & 10.40 \\
DINOv3 & 384 & 1.98 \\
DiT & 768 & 4.08 \\
Caption + text & 1,536 & 1,314.31 \\
\bottomrule
\end{tabular}}
\end{table}

The caption-mediated pipeline records approximately 1,314 milliseconds per image, compared with 10.40 for Qwen3-VL and 66.07 for BLIP-2. Since local extraction and API-based caption generation involve different computational and communication costs, these timings provide a high-level comparison of the evaluated pipelines under their respective execution conditions. The caption-mediated route nevertheless incurs substantially greater elapsed extraction time, reflecting the additional caption-generation and text-encoding stages. Monetary API expenditure and pretraining costs are not included in this comparison.

\begin{figure}[!t]
\centering

  \includegraphics[width=0.5\linewidth]{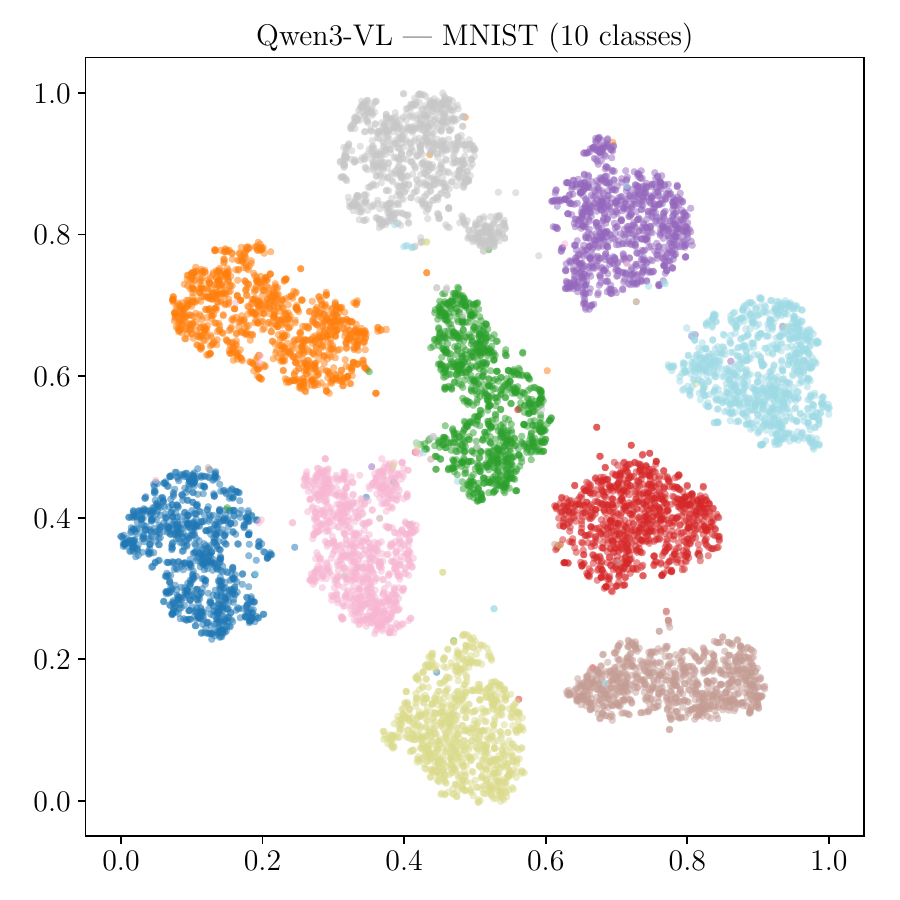}%
  \includegraphics[width=0.5\linewidth]{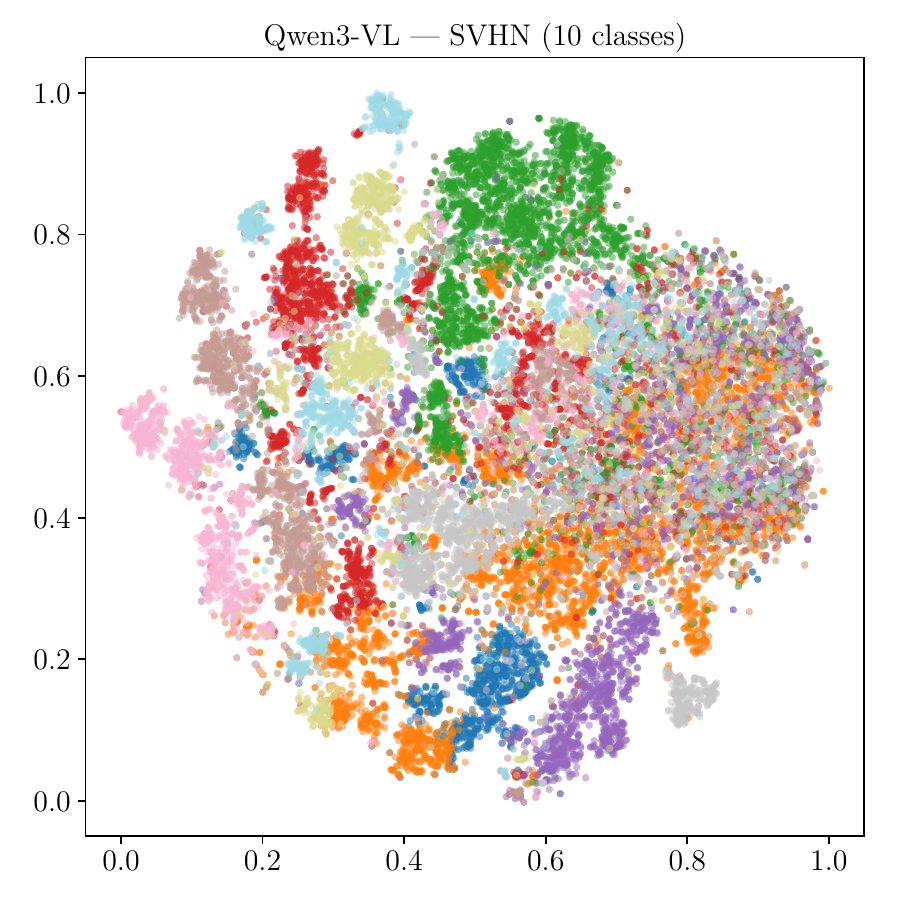}\\
  \includegraphics[width=0.5\linewidth]{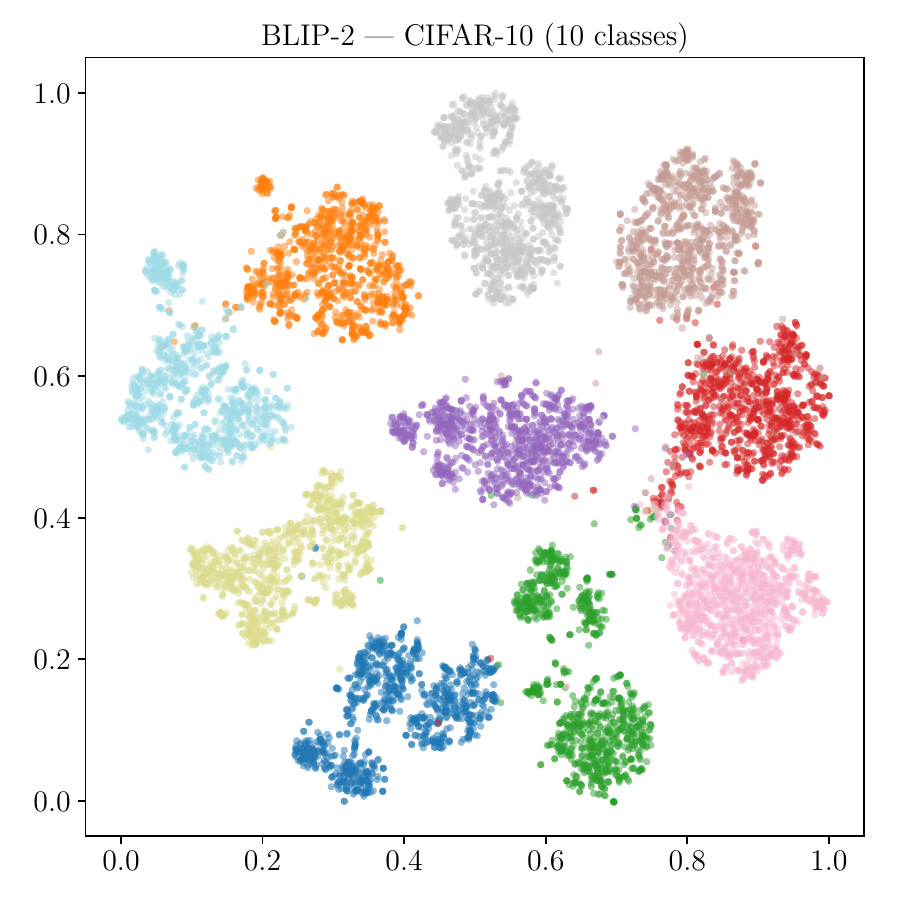}%
  \includegraphics[width=0.5\linewidth]{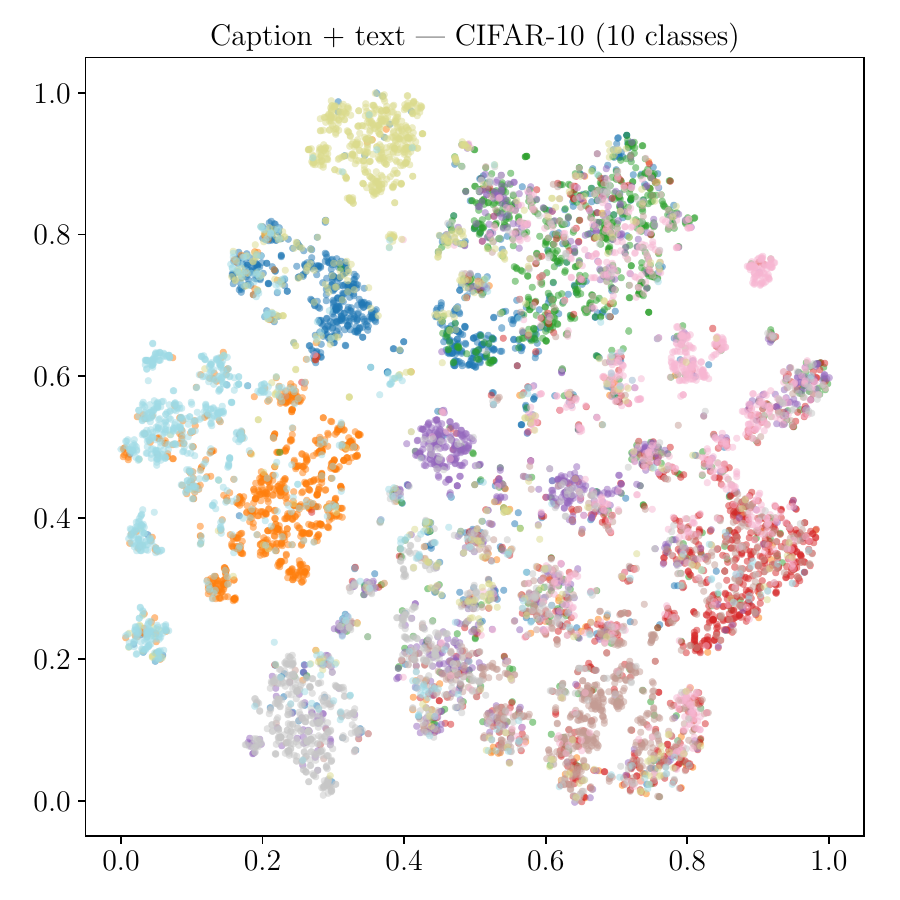}

\caption{Illustrative PCA--t-SNE projections selected retrospectively from the quantitative results. Top: Qwen3-VL on MNIST (left) and SVHN (right), contrasting domains for a fixed representation. Bottom: BLIP-2 (left) and caption-mediated embeddings (right) on CIFAR-10, contrasting representation routes for a fixed dataset. Colors denote reference classes, not predicted clusters. Panels are independently projected and rescaled; visible separation does not validate clustering in the original embedding spaces.}\label{fig:projections}
\end{figure}

Figure~\ref{fig:projections} presents four illustrative PCA--t-SNE projections selected to highlight two contrasts observed in the quantitative results. The top row holds the representation fixed, showing Qwen3-VL embeddings for MNIST (left) and SVHN (right), with corresponding AHC-Ward ARIs of 0.925 and 0.097. This comparison illustrates the difference between isolated handwritten digits and street-view digits, for which the evaluated representation recovers the reference classes with markedly different levels of agreement. The bottom row holds the dataset fixed, contrasting BLIP-2 visual embeddings (left) with caption-mediated embeddings (right) on CIFAR-10, with AHC-Ward ARIs of 0.925 and 0.273, respectively. This illustrates the stronger class agreement obtained from direct visual features in this particular comparison.

These projections provide a qualitative view of the embedding spaces rather than an additional basis for ranking representations. All reported clustering scores were computed from the original embeddings, before dimensionality reduction. The PCA--t-SNE projections and independent coordinate rescaling can alter apparent distances and separation, while the colors denote reference classes rather than predicted clusters. Accordingly, the visualizations complement the quantitative results by illustrating selected differences in representation structure.

\subsection{Synthesis of findings}
\label{sec:results:synthesis}

These results highlight the importance of evaluating complete representation--clustering pipelines across image domains. Qwen3-VL achieves the highest ARI on five datasets with AHC-Ward fixed, while BLIP-2 leads on both CIFAR datasets. However, the consistently low agreement on SVHN illustrates that strong performance on isolated handwritten digits does not necessarily extend to other digit domains. L2 normalization preserves the leading representation in each dataset but changes individual scores and performance gaps, while the choice of clustering configuration can further affect the results. AHC-Ward generally achieves the highest dataset-averaged ARI, whereas fuzzy c-means performs poorly on the original embeddings, with exploratory evidence suggesting sensitivity to dimensionality. The caption-mediated pipeline performs relatively well on isolated digits but does not consistently surpass direct visual features and incurs substantially greater extraction time.

Overall, these findings suggest that clustering suitability depends on the interaction between the extracted representation, the target dataset, and the clustering procedure, including preprocessing choices. The evaluated results favor selecting pipelines based on their behavior in the intended domain rather than relying on model characteristics or performance on a single benchmark.

\section{Limitations}

The comparison evaluates complete representation pipelines, combining pretrained checkpoints, image preprocessing, feature extraction, and pooling. Therefore, differences in clustering performance cannot be attributed to model architecture alone. The results apply to the extraction paths and clustering settings described in Section~\ref{sec:methods}; alternative layers, pooling rules, or implementation-dependent defaults may yield different outcomes. The study focuses on embeddings at their original dimensionality, without applying dimensionality reduction before clustering. Although transformations such as PCA may improve performance for some configurations, particularly fuzzy c-means, systematically evaluating them would introduce an additional experimental variable beyond the intended comparison. The L2 sensitivity analysis examines vector-length normalization with AHC-Ward fixed, but does not evaluate normalization with other algorithms or isolate the effects of pooling, extraction layer, and feature-wise scaling.

Each configuration in the baseline comparison was evaluated once, with additional raw-vector reruns serving as controls for the AHC-Ward normalization analysis. Thus, sensitivity to stochastic initialization was not estimated. Moreover, restart budgets differ between the two K-means configurations, while fuzzy c-means was not subjected to a systematic parameter-sensitivity analysis. These settings limit comparisons of initialization strategies and clustering methods under equally tuned conditions. The highest-ARI combinations were also selected retrospectively using test labels, so they represent observed maxima rather than unbiased estimates of performance under a label-independent selection procedure. Similarly, the AHC-Ward-based ranks and win counts summarize seven related benchmarks, not independent experimental repetitions. Their interpretation is descriptive, and broader evaluation would be needed to establish whether the observed rankings generalize.

All experiments use known class counts and relatively small benchmark images. Therefore, the findings do not establish performance for unknown-$K$ discovery, open-world categories, or larger real-world image collections. Possible overlap between pretraining data and public benchmarks further limits claims about generalization to unseen distributions. Dataset domain and class count also vary together, making it difficult to attribute differences in performance to either factor independently. In addition, EMNIST preprocessing does not include a separate orientation-correction ablation, leaving sensitivity to character orientation unexplored.

The caption-mediated pipeline receives dataset-specific semantic guidance that direct visual extraction does not receive. Its results thus reflect both the representation route and the prompting strategy. Since caption recognition accuracy was not assessed separately, downstream errors cannot be attributed specifically to recognition, verbalization, or text encoding. Evaluation also introduces some metric-specific considerations: weighted F1 depends on greedy centroid-based label alignment, while internal indices describe geometry in different embedding spaces. Accordingly, ARI and V-measure remain the principal measures of agreement with reference classes. Future comparisons incorporating conventional visual baselines, matched initialization budgets, and controlled prompting and pooling would help clarify which pipeline choices account for the observed differences.

\section{Conclusions}

This study compared nine pretrained representation pipelines across seven image datasets and four clustering configurations, yielding 252 dataset--representation--clustering combinations. The comparison extends the question of clustering with pretrained embeddings from text to images, considering both direct visual features and generated descriptions encoded as text. The results show variation in agreement with reference classes across image domains and partitioning procedures; no representation achieves the highest ARI in every evaluated setting.

With AHC-Ward clustering held fixed, Qwen3-VL achieved the highest ARI on five datasets, while BLIP-2 led on CIFAR-10 and CIFAR-100. Qwen3-VL also had the best mean rank in all seven datasets. This domain-dependent ordering supports the evaluation of representations in data representative of the intended application before selecting a pipeline. Despite both MNIST and SVHN containing digits, the highest ARI on SVHN was 0.097 in the baseline and 0.098 after L2 normalization with AHC-Ward. Thus, high clustering agreement on isolated handwritten digits was not accompanied by comparable agreement on street-view digits.

The paired L2 analysis preserves the highest-ARI representation in all seven datasets. However, ARI increases in 35 representation--dataset pairs, decreases in 26, and remains numerically unchanged in two. LLaVA and Qwen3-VL improve on all seven datasets. Therefore, normalization remains a pipeline choice to evaluate for the intended representation and domain; the unchanged leading representations do not imply unchanged performance differences or complete rankings.

The clustering configuration also influenced performance and representation rankings. AHC-Ward achieved the highest dataset-averaged ARI for eight of the nine representations, with random-initialized K-means leading for Qwen3-VL. Fuzzy c-means performed substantially worse overall, although DINOv3 achieved its highest mean ARI, possibly reflecting the greater suitability of its lower-dimensional embeddings for this method. The exploratory dimensionality-reduction diagnostic further suggests that preprocessing choices can substantially affect clustering outcomes. Together, these findings reinforce the importance of evaluating representations and clustering procedures jointly instead of treating their performance as independent.

The evaluated caption-mediated pipeline obtained higher agreement on isolated handwritten digits than on clothing, mixed characters, or natural objects. Even with dataset-specific semantic instructions, it did not consistently exceed direct visual representations. This finding applies to the evaluated captioning model, text encoder, and prompts. Recognition errors and the loss of visual distinctions during verbalization are plausible explanations, but require a separate assessment of caption content. Embedding dimensionality alone was also insufficient to characterize performance: Qwen3-VL and LLaVA both produced 4,096-dimensional vectors but achieved different clustering scores.

The observed rankings and dataset wins provide a descriptive basis for selecting candidate representation--clustering pipelines within the evaluated settings. Building on these findings, future studies could investigate individual methodological factors in greater depth, including clustering initialization and parameter sensitivity, alternative feature extraction and pooling strategies, and the effects of normalization and dimensionality reduction on clustering performance. Separate investigations of caption prompting, generation models, and recognition errors could further clarify the strengths and limitations of language-mediated representations. Extending these comparisons to unknown class counts and broader image collections would help assess how the observed patterns generalize beyond the present benchmarks. These directions provide opportunities for more focused analyses of the factors influencing image clustering with pretrained representations.

\section*{Declaration of competing interest}

The authors declare that they have no known competing financial interests or personal relationships that could have appeared to influence the work reported in this paper.

\section*{Declaration of generative AI and AI-assisted technologies in the manuscript preparation process}

During the preparation of this manuscript, the authors used ChatGPT (OpenAI) to support language editing, improve text flow, and obtain limited technical clarifications related to \LaTeX{} and Python code, as well as high-level suggestions on section organization and draft wording for the abstract and conclusions. All scientific ideas, model design choices, experimental rationale, data analysis, interpretation of results, and conclusions were developed by the authors. All content generated with AI assistance was carefully reviewed, edited, and, where appropriate, substantially modified by the authors, who take full responsibility for the final content of the published article.

\section*{Acknowledgments}

This work was supported by the Funda\c c\~ao para a Ci\^encia e a Tecnologia (Portugal, \url{https://ror.org/00snfqn58}) under grants/projects 
2023.15441.TENURE.051/CP00003/CT00029 (João P. Matos-Carvalho),
CEECINST/00002/2021/CP2788\-/\-CT0001 (Nuno Fachada, \url{https://doi.org/10.54499/CEECINST/00002/2021/CP2788/CT0001}), 
UID/00408/2025 (LASIGE R\&D Unit, \url{https://doi.org/10.54499/UID/00408/2025}), 
UID/PRR/00408/2025 (LASIGE R\&D Unit, \url{https://doi.org/10.54499/UID/PRR/00408/2025}),
UID/06486/2025 (INESC INOV R\&D Unit, \url{https://doi.org/10.54499/UID/06486/2025}),
UID/PRR/06486/2025 (INESC INOV R\&D Unit, \url{https://doi.org/10.54499/UID/PRR/06486/2025}),
and
UID/PRR2/06486/2025 (INESC INOV R\&D Unit, \url{https://doi.org/10.54499/UID/PRR2/06486/2025}).

\bibliographystyle{elsarticle-num}

\clearpage
\appendix

\section{Complementary V-measure results for AHC-Ward clustering}
\label{sec:apx:vmeasure}

This appendix provides complementary V-measure results for AHC-Ward clustering across the nine representations and seven datasets, using the original embeddings (Table~\ref{tab:ward-vmeasure}) and their L2-normalized counterparts (Table~\ref{tab:l2-v_measure}). These correspond to the ARI results in Tables~\ref{tab:ward-ari} and~\ref{tab:l2-ari}, respectively.

 % tables/ward_vmeasure
\begin{table}[!h]
\centering
{\scriptsize{}%
\caption{V-measure with a fixed AHC-Ward clustering algorithm. Bold values indicate the best representation for each dataset.}
\label{tab:ward-vmeasure}
\begin{tabular}{lrrrrrrr}
\toprule
Representation & MNIST & Fashion & EMNIST & USPS & SVHN & CIFAR-10 & CIFAR-100 \\
\midrule
CLIP & 0.630 & 0.675 & 0.557 & 0.578 & 0.023 & 0.641 & 0.552 \\
SigLIP2 & 0.672 & 0.688 & 0.611 & 0.752 & 0.025 & 0.705 & 0.657 \\
BLIP & 0.757 & 0.659 & 0.601 & 0.713 & 0.018 & 0.605 & 0.548 \\
BLIP-2 & 0.726 & \textbf{0.720} & 0.603 & 0.740 & 0.084 & \textbf{0.938} & \textbf{0.822} \\
LLaVA & 0.348 & 0.413 & 0.357 & 0.398 & 0.022 & 0.494 & 0.530 \\
Qwen3-VL & \textbf{0.934} & 0.698 & \textbf{0.687} & \textbf{0.824} & \textbf{0.158} & 0.860 & 0.781 \\
DINOv3 & 0.646 & 0.679 & 0.565 & 0.677 & 0.023 & 0.726 & 0.626 \\
DiT & 0.354 & 0.333 & 0.309 & 0.297 & 0.004 & 0.061 & 0.190 \\
Caption + text & 0.809 & 0.312 & 0.295 & 0.729 & 0.039 & 0.382 & 0.411 \\
\bottomrule
\end{tabular}}
\end{table}

 % tables/l2_v_measure
\begin{table}[!ht]
\centering
{\scriptsize{}%
\caption{V-measure with AHC-Ward after per-image L2 normalization, retaining the original embedding dimensions. Bold marks the highest value in each dataset before rounding.}
\label{tab:l2-v_measure}
\begin{tabular}{lrrrrrrr}
\toprule
Representation & MNIST & Fashion & EMNIST & USPS & SVHN & CIFAR-10 & CIFAR-100 \\
\midrule
CLIP & 0.622 & 0.656 & 0.561 & 0.629 & 0.021 & 0.608 & 0.559 \\
SigLIP2 & 0.704 & 0.679 & 0.604 & 0.732 & 0.045 & 0.726 & 0.654 \\
BLIP & 0.780 & 0.651 & 0.590 & 0.685 & 0.024 & 0.612 & 0.554 \\
BLIP-2 & 0.685 & 0.714 & 0.605 & 0.770 & 0.079 & \textbf{0.928} & \textbf{0.823} \\
LLaVA & 0.514 & 0.627 & 0.431 & 0.508 & 0.020 & 0.564 & 0.562 \\
Qwen3-VL & \textbf{0.939} & \textbf{0.718} & \textbf{0.688} & \textbf{0.872} & \textbf{0.183} & 0.910 & 0.786 \\
DINOv3 & 0.683 & 0.702 & 0.572 & 0.650 & 0.019 & 0.721 & 0.632 \\
DiT & 0.322 & 0.372 & 0.317 & 0.298 & 0.003 & 0.064 & 0.188 \\
Caption + text & 0.809 & 0.312 & 0.298 & 0.729 & 0.033 & 0.381 & 0.411 \\
\bottomrule
\end{tabular}}
\end{table}

\clearpage

\clearpage

\section{Detailed results for the effect of clustering configuration experiment}
\label{sec:apx:cluconf}

Tables~\ref{tab:full-mnist},~\ref{tab:full-fashion_mnist},~\ref{tab:full-emnist},~\ref{tab:full-usps},~\ref{tab:full-svhn},~\ref{tab:full-cifar10} and~\ref{tab:full-cifar100} present the ARI, V-measure, weighted F1 (greedy), and silhouette results for all clustering configurations for the MNIST, Fashion-MNIST, EMNIST, USPS, SVHN, CIFAR-10, and CIFAR-100 datasets, respectively.

These tables complement the aggregate results presented in Section~\ref{sec:results:cluconf}.

 % tables/full_mnist
\begin{table}[!ht]
\centering
{\scriptsize{}%
\caption{Detailed results for MNIST for all clustering configurations. Bold marks column maxima before rounding for ARI, V-measure, and F1 (greedy centroid mapping). Silhouette is not compared across feature spaces.}
\label{tab:full-mnist}
\begin{tabular}{llrrrr}
\toprule
Representation & Algorithm & ARI & V-measure & F1 (greedy) & Silhouette \\
\midrule
CLIP & K-means++ & 0.436 & 0.520 & 0.656 & 0.111 \\
     & K-means (random) & 0.373 & 0.498 & 0.591 & 0.118 \\
     & AHC-Ward & 0.511 & 0.630 & 0.636 & 0.088 \\
     & Fuzzy c-means & 0.062 & 0.126 & 0.049 & 0.157 \\
\addlinespace
SigLIP2 & K-means++ & 0.488 & 0.616 & 0.697 & 0.167 \\
        & K-means (random) & 0.488 & 0.623 & 0.643 & 0.165 \\
        & AHC-Ward & 0.479 & 0.672 & 0.630 & 0.149 \\
        & Fuzzy c-means & 0.071 & 0.145 & 0.038 & 0.163 \\
\addlinespace
BLIP & K-means++ & 0.580 & 0.696 & 0.701 & 0.144 \\
     & K-means (random) & 0.638 & 0.716 & 0.769 & 0.151 \\
     & AHC-Ward & 0.614 & 0.757 & 0.719 & 0.128 \\
     & Fuzzy c-means & 0.091 & 0.189 & 0.048 & 0.131 \\
\addlinespace
BLIP-2 & K-means++ & 0.425 & 0.553 & 0.536 & 0.141 \\
       & K-means (random) & 0.431 & 0.559 & 0.628 & 0.146 \\
       & AHC-Ward & 0.588 & 0.726 & 0.765 & 0.122 \\
       & Fuzzy c-means & 0.067 & 0.134 & 0.047 & 0.143 \\
\addlinespace
LLaVA & K-means++ & 0.111 & 0.221 & 0.214 & 0.101 \\
      & K-means (random) & 0.139 & 0.258 & 0.311 & 0.105 \\
      & AHC-Ward & 0.176 & 0.348 & 0.366 & 0.079 \\
      & Fuzzy c-means & 0.025 & 0.040 & 0.101 & -0.009 \\
\addlinespace
Qwen3-VL & K-means++ & \textbf{0.941} & 0.933 & 0.973 & 0.272 \\
         & K-means (random) & 0.941 & 0.933 & \textbf{0.973} & 0.272 \\
         & AHC-Ward & 0.925 & \textbf{0.934} & 0.968 & 0.262 \\
         & Fuzzy c-means & 0.092 & 0.206 & 0.078 & 0.193 \\
\addlinespace
DINOv3 & K-means++ & 0.444 & 0.531 & 0.575 & 0.118 \\
       & K-means (random) & 0.398 & 0.526 & 0.554 & 0.117 \\
       & AHC-Ward & 0.468 & 0.646 & 0.612 & 0.089 \\
       & Fuzzy c-means & 0.191 & 0.326 & 0.091 & -0.000 \\
\addlinespace
DiT & K-means++ & 0.163 & 0.291 & 0.313 & 0.120 \\
    & K-means (random) & 0.166 & 0.287 & 0.287 & 0.121 \\
    & AHC-Ward & 0.208 & 0.354 & 0.278 & 0.083 \\
    & Fuzzy c-means & 0.119 & 0.189 & 0.213 & 0.029 \\
\addlinespace
Caption + text & K-means++ & 0.756 & 0.800 & 0.811 & 0.234 \\
               & K-means (random) & 0.770 & 0.818 & 0.858 & 0.236 \\
               & AHC-Ward & 0.793 & 0.809 & 0.904 & 0.245 \\
               & Fuzzy c-means & 0.078 & 0.170 & 0.058 & -0.021 \\
\bottomrule
\end{tabular}}
\end{table}
\clearpage
 % tables/full_fashion_mnist
\begin{table}[!ht]
\centering
{\scriptsize{}%
\caption{Detailed results for Fashion-MNIST for all clustering configurations. Bold marks column maxima before rounding for ARI, V-measure, and F1 (greedy centroid mapping). Silhouette is not compared across feature spaces.}
\label{tab:full-fashion_mnist}
\begin{tabular}{llrrrr}
\toprule
Representation & Algorithm & ARI & V-measure & F1 (greedy) & Silhouette \\
\midrule
CLIP & K-means++ & 0.481 & 0.638 & 0.527 & 0.182 \\
    & K-means (random) & 0.515 & 0.636 & 0.644 & 0.159 \\
    & AHC-Ward & 0.534 & 0.675 & 0.651 & 0.160 \\
    & Fuzzy c-means & 0.145 & 0.319 & 0.051 & 0.180 \\
\addlinespace
SigLIP2 & K-means++ & 0.534 & 0.668 & 0.699 & 0.167 \\
    & K-means (random) & 0.515 & 0.660 & 0.687 & 0.166 \\
    & AHC-Ward & 0.529 & 0.688 & 0.651 & 0.162 \\
    & Fuzzy c-means & 0.128 & 0.287 & 0.049 & 0.170 \\
\addlinespace
BLIP & K-means++ & 0.500 & 0.635 & 0.676 & 0.168 \\
    & K-means (random) & 0.477 & 0.637 & 0.602 & 0.193 \\
    & AHC-Ward & 0.515 & 0.659 & 0.617 & 0.171 \\
    & Fuzzy c-means & 0.132 & 0.282 & 0.053 & 0.171 \\
\addlinespace
BLIP-2 & K-means++ & 0.596 & 0.722 & 0.717 & 0.131 \\
    & K-means (random) & 0.550 & 0.701 & 0.674 & 0.131 \\
    & AHC-Ward & 0.526 & 0.720 & 0.605 & 0.127 \\
    & Fuzzy c-means & 0.171 & 0.391 & 0.065 & 0.126 \\
\addlinespace
LLaVA & K-means++ & 0.245 & 0.411 & 0.350 & 0.139 \\
    & K-means (random) & 0.251 & 0.404 & 0.399 & 0.141 \\
    & AHC-Ward & 0.237 & 0.413 & 0.342 & 0.109 \\
    & Fuzzy c-means & 0.110 & 0.203 & 0.167 & 0.038 \\
\addlinespace
Qwen3-VL & K-means++ & 0.575 & 0.708 & 0.684 & 0.165 \\
    & K-means (random) & \textbf{0.630} & \textbf{0.728} & \textbf{0.751} & 0.162 \\
    & AHC-Ward & 0.573 & 0.698 & 0.705 & 0.146 \\
    & Fuzzy c-means & 0.155 & 0.327 & 0.060 & 0.012 \\
\addlinespace
DINOv3 & K-means++ & 0.523 & 0.672 & 0.642 & 0.149 \\
    & K-means (random) & 0.477 & 0.676 & 0.578 & 0.161 \\
    & AHC-Ward & 0.485 & 0.679 & 0.606 & 0.151 \\
    & Fuzzy c-means & 0.232 & 0.474 & 0.208 & 0.154 \\
\addlinespace
DiT & K-means++ & 0.211 & 0.327 & 0.300 & 0.201 \\
    & K-means (random) & 0.211 & 0.327 & 0.300 & 0.201 \\
    & AHC-Ward & 0.238 & 0.333 & 0.243 & 0.149 \\
    & Fuzzy c-means & 0.182 & 0.258 & 0.216 & 0.138 \\
\addlinespace
Caption + text & K-means++ & 0.206 & 0.312 & 0.378 & 0.141 \\
    & K-means (random) & 0.201 & 0.319 & 0.372 & 0.152 \\
    & AHC-Ward & 0.193 & 0.312 & 0.391 & 0.129 \\
    & Fuzzy c-means & 0.112 & 0.232 & 0.046 & 0.189 \\
\bottomrule
\end{tabular}}
\end{table}
\clearpage
 % tables/full_emnist
\begin{table}[!ht]
\centering
{\scriptsize{}%
\caption{Detailed results for EMNIST for all clustering configurations. Bold marks column maxima before rounding for ARI, V-measure, and F1 (greedy centroid mapping). Silhouette is not compared across feature spaces.}
\label{tab:full-emnist}
\begin{tabular}{llrrrr}
\toprule
Representation & Algorithm & ARI & V-measure & F1 (greedy) & Silhouette \\
\midrule
CLIP & K-means++ & 0.228 & 0.497 & 0.365 & 0.068 \\
    & K-means (random) & 0.237 & 0.498 & 0.409 & 0.073 \\
    & AHC-Ward & 0.283 & 0.557 & 0.394 & 0.041 \\
    & Fuzzy c-means & 0.011 & 0.069 & 0.002 & -0.076 \\
\addlinespace
SigLIP2 & K-means++ & 0.280 & 0.545 & 0.448 & 0.087 \\
    & K-means (random) & 0.282 & 0.549 & 0.451 & 0.091 \\
    & AHC-Ward & 0.332 & 0.611 & 0.486 & 0.061 \\
    & Fuzzy c-means & 0.022 & 0.144 & 0.006 & -0.027 \\
\addlinespace
BLIP & K-means++ & 0.313 & 0.573 & 0.481 & 0.090 \\
    & K-means (random) & 0.302 & 0.562 & 0.467 & 0.086 \\
    & AHC-Ward & 0.324 & 0.601 & 0.471 & 0.059 \\
    & Fuzzy c-means & 0.012 & 0.077 & 0.002 & -0.026 \\
\addlinespace
BLIP-2 & K-means++ & 0.301 & 0.551 & 0.458 & 0.084 \\
    & K-means (random) & 0.307 & 0.556 & 0.471 & 0.088 \\
    & AHC-Ward & 0.327 & 0.603 & 0.485 & 0.051 \\
    & Fuzzy c-means & 0.012 & 0.072 & 0.002 & -0.055 \\
\addlinespace
LLaVA & K-means++ & 0.099 & 0.323 & 0.188 & 0.064 \\
    & K-means (random) & 0.100 & 0.317 & 0.189 & 0.065 \\
    & AHC-Ward & 0.106 & 0.357 & 0.201 & 0.027 \\
    & Fuzzy c-means & 0.019 & 0.085 & 0.024 & -0.030 \\
\addlinespace
Qwen3-VL & K-means++ & 0.394 & 0.645 & 0.536 & 0.118 \\
    & K-means (random) & 0.422 & 0.657 & \textbf{0.568} & 0.119 \\
    & AHC-Ward & \textbf{0.434} & \textbf{0.687} & 0.554 & 0.093 \\
    & Fuzzy c-means & 0.017 & 0.118 & 0.005 & -0.035 \\
\addlinespace
DINOv3 & K-means++ & 0.265 & 0.515 & 0.405 & 0.088 \\
    & K-means (random) & 0.270 & 0.521 & 0.429 & 0.085 \\
    & AHC-Ward & 0.303 & 0.565 & 0.451 & 0.047 \\
    & Fuzzy c-means & 0.028 & 0.173 & 0.006 & -0.121 \\
\addlinespace
DiT & K-means++ & 0.083 & 0.299 & 0.160 & 0.078 \\
    & K-means (random) & 0.080 & 0.298 & 0.153 & 0.079 \\
    & AHC-Ward & 0.083 & 0.309 & 0.152 & 0.027 \\
    & Fuzzy c-means & 0.027 & 0.131 & 0.033 & -0.062 \\
\addlinespace
Caption + text & K-means++ & 0.083 & 0.281 & 0.181 & 0.112 \\
    & K-means (random) & 0.088 & 0.291 & 0.200 & 0.120 \\
    & AHC-Ward & 0.104 & 0.295 & 0.199 & 0.109 \\
    & Fuzzy c-means & 0.003 & 0.049 & 0.006 & -0.060 \\
\bottomrule
\end{tabular}}
\end{table}
\clearpage
 % tables/full_usps
\begin{table}[!ht]
\centering
{\scriptsize{}%
\caption{Detailed results for USPS for all clustering configurations. Bold marks column maxima before rounding for ARI, V-measure, and F1 (greedy centroid mapping). Silhouette is not compared across feature spaces.}
\label{tab:full-usps}
\begin{tabular}{llrrrr}
\toprule
Representation & Algorithm & ARI & V-measure & F1 (greedy) & Silhouette \\
\midrule
CLIP & K-means++ & 0.390 & 0.527 & 0.576 & 0.169 \\
    & K-means (random) & 0.354 & 0.530 & 0.482 & 0.172 \\
    & AHC-Ward & 0.426 & 0.578 & 0.542 & 0.147 \\
    & Fuzzy c-means & 0.104 & 0.159 & 0.037 & 0.162 \\
\addlinespace
SigLIP2 & K-means++ & 0.472 & 0.646 & 0.584 & 0.141 \\
    & K-means (random) & 0.547 & 0.667 & 0.695 & 0.196 \\
    & AHC-Ward & 0.687 & 0.752 & 0.774 & 0.207 \\
    & Fuzzy c-means & 0.139 & 0.243 & 0.035 & 0.184 \\
\addlinespace
BLIP & K-means++ & 0.543 & 0.632 & 0.689 & 0.197 \\
    & K-means (random) & 0.533 & 0.629 & 0.683 & 0.194 \\
    & AHC-Ward & 0.579 & 0.713 & 0.637 & 0.165 \\
    & Fuzzy c-means & 0.147 & 0.259 & 0.045 & 0.152 \\
\addlinespace
BLIP-2 & K-means++ & 0.507 & 0.637 & 0.591 & 0.166 \\
    & K-means (random) & 0.648 & 0.729 & 0.753 & 0.171 \\
    & AHC-Ward & 0.685 & 0.740 & 0.800 & 0.166 \\
    & Fuzzy c-means & 0.120 & 0.200 & 0.031 & 0.137 \\
\addlinespace
LLaVA & K-means++ & 0.285 & 0.362 & 0.428 & 0.160 \\
    & K-means (random) & 0.255 & 0.372 & 0.370 & 0.154 \\
    & AHC-Ward & 0.266 & 0.398 & 0.387 & 0.119 \\
    & Fuzzy c-means & 0.165 & 0.225 & 0.262 & 0.087 \\
\addlinespace
Qwen3-VL & K-means++ & 0.582 & 0.749 & 0.664 & 0.231 \\
    & K-means (random) & \textbf{0.759} & \textbf{0.834} & \textbf{0.813} & 0.245 \\
    & AHC-Ward & 0.717 & 0.824 & 0.798 & 0.234 \\
    & Fuzzy c-means & 0.147 & 0.268 & 0.130 & 0.186 \\
\addlinespace
DINOv3 & K-means++ & 0.426 & 0.562 & 0.521 & 0.172 \\
    & K-means (random) & 0.463 & 0.577 & 0.610 & 0.172 \\
    & AHC-Ward & 0.612 & 0.677 & 0.720 & 0.152 \\
    & Fuzzy c-means & 0.109 & 0.186 & 0.041 & 0.035 \\
\addlinespace
DiT & K-means++ & 0.149 & 0.268 & 0.289 & 0.173 \\
    & K-means (random) & 0.158 & 0.281 & 0.266 & 0.181 \\
    & AHC-Ward & 0.153 & 0.297 & 0.252 & 0.143 \\
    & Fuzzy c-means & 0.233 & 0.268 & 0.278 & 0.125 \\
\addlinespace
Caption + text & K-means++ & 0.644 & 0.749 & 0.728 & 0.192 \\
    & K-means (random) & 0.645 & 0.725 & 0.735 & 0.195 \\
    & AHC-Ward & 0.694 & 0.729 & 0.793 & 0.214 \\
    & Fuzzy c-means & 0.040 & 0.127 & 0.048 & -0.043 \\
\bottomrule
\end{tabular}}
\end{table}
\clearpage
 % tables/full_svhn
\begin{table}[!ht]
\centering
{\scriptsize{}%
\caption{Detailed results for SVHN for all clustering configurations. Bold marks column maxima before rounding for ARI, V-measure, and F1 (greedy centroid mapping). Silhouette is not compared across feature spaces.}
\label{tab:full-svhn}
\begin{tabular}{llrrrr}
\toprule
Representation & Algorithm & ARI & V-measure & F1 (greedy) & Silhouette \\
\midrule
CLIP & K-means++ & 0.016 & 0.023 & 0.138 & 0.042 \\
    & K-means (random) & 0.017 & 0.024 & 0.143 & 0.043 \\
    & AHC-Ward & 0.013 & 0.023 & 0.143 & 0.007 \\
    & Fuzzy c-means & 0.009 & 0.009 & 0.064 & 0.033 \\
\addlinespace
SigLIP2 & K-means++ & 0.014 & 0.024 & 0.138 & 0.059 \\
    & K-means (random) & 0.014 & 0.025 & 0.138 & 0.059 \\
    & AHC-Ward & 0.012 & 0.025 & 0.111 & 0.025 \\
    & Fuzzy c-means & 0.019 & 0.015 & 0.078 & 0.083 \\
\addlinespace
BLIP & K-means++ & 0.017 & 0.027 & 0.150 & 0.054 \\
    & K-means (random) & 0.012 & 0.020 & 0.148 & 0.052 \\
    & AHC-Ward & 0.010 & 0.018 & 0.118 & 0.014 \\
    & Fuzzy c-means & 0.006 & 0.008 & 0.036 & 0.182 \\
\addlinespace
BLIP-2 & K-means++ & 0.050 & 0.083 & 0.182 & 0.064 \\
    & K-means (random) & 0.050 & 0.082 & 0.182 & 0.064 \\
    & AHC-Ward & 0.033 & 0.084 & 0.197 & 0.029 \\
    & Fuzzy c-means & 0.005 & 0.015 & 0.077 & -0.002 \\
\addlinespace
LLaVA & K-means++ & 0.011 & 0.023 & 0.135 & 0.076 \\
    & K-means (random) & 0.012 & 0.025 & 0.133 & 0.076 \\
    & AHC-Ward & 0.005 & 0.022 & 0.108 & 0.031 \\
    & Fuzzy c-means & 0.001 & 0.011 & 0.111 & -0.028 \\
\addlinespace
Qwen3-VL & K-means++ & 0.083 & 0.124 & 0.191 & 0.056 \\
    & K-means (random) & 0.069 & 0.093 & 0.138 & 0.055 \\
    & AHC-Ward & \textbf{0.097} & \textbf{0.158} & \textbf{0.217} & 0.018 \\
    & Fuzzy c-means & -0.001 & 0.015 & 0.052 & 0.078 \\
\addlinespace
DINOv3 & K-means++ & 0.018 & 0.034 & 0.163 & 0.058 \\
    & K-means (random) & 0.017 & 0.035 & 0.148 & 0.058 \\
    & AHC-Ward & 0.010 & 0.023 & 0.131 & 0.027 \\
    & Fuzzy c-means & 0.003 & 0.008 & 0.037 & -0.066 \\
\addlinespace
DiT & K-means++ & 0.003 & 0.005 & 0.117 & 0.230 \\
    & K-means (random) & 0.002 & 0.004 & 0.103 & 0.224 \\
    & AHC-Ward & 0.003 & 0.004 & 0.118 & 0.176 \\
    & Fuzzy c-means & 0.002 & 0.004 & 0.114 & 0.204 \\
\addlinespace
Caption + text & K-means++ & 0.005 & 0.009 & 0.125 & 0.097 \\
    & K-means (random) & 0.004 & 0.009 & 0.120 & 0.102 \\
    & AHC-Ward & 0.015 & 0.039 & 0.138 & 0.075 \\
    & Fuzzy c-means & 0.001 & 0.001 & 0.069 & 0.100 \\
\bottomrule
\end{tabular}}
\end{table}
\clearpage
 % tables/full_cifar10
\begin{table}[!ht]
\centering
{\scriptsize{}%
\caption{Detailed results for CIFAR-10 for all clustering configurations. Bold marks column maxima before rounding for ARI, V-measure, and F1 (greedy centroid mapping). Silhouette is not compared across feature spaces.}
\label{tab:full-cifar10}
\begin{tabular}{llrrrr}
\toprule
Representation & Algorithm & ARI & V-measure & F1 (greedy) & Silhouette \\
\midrule
CLIP & K-means++ & 0.457 & 0.605 & 0.705 & 0.098 \\
    & K-means (random) & 0.465 & 0.605 & 0.704 & 0.101 \\
    & AHC-Ward & 0.508 & 0.641 & 0.689 & 0.075 \\
    & Fuzzy c-means & 0.095 & 0.190 & 0.045 & 0.106 \\
\addlinespace
SigLIP2 & K-means++ & 0.617 & 0.731 & 0.805 & 0.142 \\
    & K-means (random) & 0.632 & 0.743 & 0.786 & 0.144 \\
    & AHC-Ward & 0.575 & 0.705 & 0.708 & 0.120 \\
    & Fuzzy c-means & 0.178 & 0.408 & 0.066 & -0.000 \\
\addlinespace
BLIP & K-means++ & 0.484 & 0.597 & 0.635 & 0.098 \\
    & K-means (random) & 0.508 & 0.631 & 0.717 & 0.102 \\
    & AHC-Ward & 0.453 & 0.605 & 0.708 & 0.082 \\
    & Fuzzy c-means & 0.153 & 0.327 & 0.067 & -0.044 \\
\addlinespace
BLIP-2 & K-means++ & 0.866 & 0.932 & 0.851 & 0.159 \\
    & K-means (random) & \textbf{0.970} & \textbf{0.964} & \textbf{0.986} & 0.168 \\
    & AHC-Ward & 0.925 & 0.938 & 0.967 & 0.164 \\
    & Fuzzy c-means & 0.177 & 0.412 & 0.068 & -0.024 \\
\addlinespace
LLaVA & K-means++ & 0.280 & 0.443 & 0.423 & 0.091 \\
    & K-means (random) & 0.276 & 0.427 & 0.423 & 0.085 \\
    & AHC-Ward & 0.340 & 0.494 & 0.479 & 0.066 \\
    & Fuzzy c-means & 0.048 & 0.099 & 0.078 & 0.089 \\
\addlinespace
Qwen3-VL & K-means++ & 0.751 & 0.854 & 0.770 & 0.098 \\
    & K-means (random) & 0.798 & 0.864 & 0.886 & 0.102 \\
    & AHC-Ward & 0.777 & 0.860 & 0.806 & 0.093 \\
    & Fuzzy c-means & 0.167 & 0.365 & 0.056 & -0.064 \\
\addlinespace
DINOv3 & K-means++ & 0.656 & 0.728 & 0.798 & 0.076 \\
    & K-means (random) & 0.696 & 0.758 & 0.827 & 0.079 \\
    & AHC-Ward & 0.655 & 0.726 & 0.787 & 0.066 \\
    & Fuzzy c-means & 0.157 & 0.334 & 0.060 & 0.007 \\
\addlinespace
DiT & K-means++ & 0.035 & 0.073 & 0.137 & 0.142 \\
    & K-means (random) & 0.034 & 0.073 & 0.137 & 0.142 \\
    & AHC-Ward & 0.025 & 0.061 & 0.130 & 0.097 \\
    & Fuzzy c-means & 0.029 & 0.053 & 0.153 & 0.020 \\
\addlinespace
Caption + text & K-means++ & 0.233 & 0.351 & 0.410 & 0.092 \\
    & K-means (random) & 0.263 & 0.387 & 0.520 & 0.119 \\
    & AHC-Ward & 0.273 & 0.382 & 0.467 & 0.113 \\
    & Fuzzy c-means & 0.136 & 0.263 & 0.088 & -0.018 \\
\bottomrule
\end{tabular}}
\end{table}
\clearpage
 % tables/full_cifar100
\begin{table}[!ht]
\centering
{\scriptsize{}%
\caption{Detailed results for CIFAR-100 for all clustering configurations. Bold marks column maxima before rounding for ARI, V-measure, and F1 (greedy centroid mapping). Silhouette is not compared across feature spaces.}
\label{tab:full-cifar100}
\begin{tabular}{llrrrr}
\toprule
Representation & Algorithm & ARI & V-measure & F1 (greedy) & Silhouette \\
\midrule
CLIP & K-means++ & 0.186 & 0.536 & 0.347 & 0.050 \\
    & K-means (random) & 0.198 & 0.538 & 0.361 & 0.050 \\
    & AHC-Ward & 0.186 & 0.552 & 0.351 & 0.032 \\
    & Fuzzy c-means & 0.006 & 0.072 & 0.001 & -0.029 \\
\addlinespace
SigLIP2 & K-means++ & 0.297 & 0.634 & 0.461 & 0.075 \\
    & K-means (random) & 0.318 & 0.641 & 0.486 & 0.079 \\
    & AHC-Ward & 0.311 & 0.657 & 0.512 & 0.065 \\
    & Fuzzy c-means & 0.031 & 0.255 & 0.004 & -0.082 \\
\addlinespace
BLIP & K-means++ & 0.212 & 0.545 & 0.370 & 0.053 \\
    & K-means (random) & 0.228 & 0.553 & 0.375 & 0.057 \\
    & AHC-Ward & 0.216 & 0.548 & 0.373 & 0.038 \\
    & Fuzzy c-means & 0.020 & 0.194 & 0.003 & -0.075 \\
\addlinespace
BLIP-2 & K-means++ & 0.573 & 0.817 & 0.684 & 0.107 \\
    & K-means (random) & 0.587 & 0.821 & 0.711 & 0.113 \\
    & AHC-Ward & \textbf{0.591} & \textbf{0.822} & \textbf{0.739} & 0.103 \\
    & Fuzzy c-means & 0.017 & 0.225 & 0.001 & -0.090 \\
\addlinespace
LLaVA & K-means++ & 0.161 & 0.508 & 0.287 & 0.053 \\
    & K-means (random) & 0.160 & 0.504 & 0.260 & 0.057 \\
    & AHC-Ward & 0.168 & 0.530 & 0.276 & 0.030 \\
    & Fuzzy c-means & 0.004 & 0.083 & 0.004 & -0.042 \\
\addlinespace
Qwen3-VL & K-means++ & 0.498 & 0.775 & 0.627 & 0.092 \\
    & K-means (random) & 0.494 & 0.775 & 0.628 & 0.092 \\
    & AHC-Ward & 0.475 & 0.781 & 0.666 & 0.081 \\
    & Fuzzy c-means & 0.020 & 0.235 & 0.002 & -0.103 \\
\addlinespace
DINOv3 & K-means++ & 0.300 & 0.624 & 0.426 & 0.055 \\
    & K-means (random) & 0.322 & 0.637 & 0.453 & 0.057 \\
    & AHC-Ward & 0.287 & 0.626 & 0.436 & 0.038 \\
    & Fuzzy c-means & 0.015 & 0.195 & 0.002 & -0.075 \\
\addlinespace
DiT & K-means++ & 0.012 & 0.194 & 0.027 & 0.086 \\
    & K-means (random) & 0.012 & 0.194 & 0.025 & 0.083 \\
    & AHC-Ward & 0.012 & 0.190 & 0.024 & 0.040 \\
    & Fuzzy c-means & 0.008 & 0.122 & 0.011 & -0.131 \\
\addlinespace
Caption + text & K-means++ & 0.116 & 0.411 & 0.228 & 0.081 \\
    & K-means (random) & 0.116 & 0.412 & 0.218 & 0.084 \\
    & AHC-Ward & 0.111 & 0.411 & 0.229 & 0.076 \\
    & Fuzzy c-means & 0.022 & 0.173 & 0.003 & -0.030 \\
\bottomrule
\end{tabular}}
\end{table}

\end{document}